\documentclass[sigconf,authorversion,nonacm,screen]{acmart}

\AtBeginDocument{%
  \providecommand\BibTeX{{%
    \normalfont B\kern-0.5em{\scshape i\kern-0.25em b}\kern-0.8em\TeX}}}

\setcopyright{acmcopyright}
\copyrightyear{2022}
\acmYear{2022}
\acmDOI{XXXXXXX.XXXXXXX}

\acmConference[KDD '22]{KDD '22: ACM SIGKDD Conference on Knowledge Discovery and Data Mining}{August 14--18, 2022}{Washington DC}
\acmBooktitle{KDD '22: ACM SIGKDD Conference on Knowledge Discovery and Data Mining, August 14--18, 2022, Washington DC}

\acmPrice{15.00}
\acmISBN{978-1-4503-XXXX-X/18/06}

\usepackage[linesnumbered,ruled,vlined]{algorithm2e}
\usepackage{todonotes}
\usepackage{graphicx}
\usepackage{subfigure}
\usepackage{multirow}
\usepackage{booktabs}
\usepackage{bbm}

\begin{document}

\title[What Makes Good Contrastive Learning on Small-Scale Wearable-based Tasks?]{What Makes Good Contrastive Learning on Small-Scale Wearable-based Tasks?}






\author{Hangwei Qian}
\email{hangwei.qian@ntu.edu.sg}
\affiliation{
  \institution{Nanyang Technological University}
  \country{Singapore}}
  
\author{Tian Tian}
\email{tian.tian@ntu.edu.sg}
\affiliation{
  \institution{Nanyang Technological University}
  \country{Singapore}}

\author{Chunyan Miao}
\email{ascymiao@ntu.edu.sg}
\affiliation{
  \institution{Nanyang Technological University}
  \country{Singapore}}


\begin{abstract}

Self-supervised learning establishes a new paradigm of learning representations with much fewer or even no label annotations. Recently there has been remarkable progress on large-scale contrastive learning models which require substantial computing resources, yet such models are not practically optimal for small-scale tasks. To fill the gap, we aim to study contrastive learning on the wearable-based activity recognition task. Specifically, we conduct an in-depth study of contrastive learning from both algorithmic-level and task-level perspectives. For algorithmic-level analysis, we decompose contrastive models into several key components and conduct rigorous experimental evaluations to better understand the efficacy and rationale behind contrastive learning. More importantly, for task-level analysis, we show that the wearable-based signals bring unique challenges and opportunities to existing contrastive models, which cannot be readily solved by existing algorithms. Our thorough empirical studies suggest important practices and shed light on future research challenges. In the meantime, this paper presents an open-source PyTorch library \texttt{CL-HAR}, which can serve as a practical tool for researchers\footnote{\texttt{CL-HAR} is available at \url{https://github.com/Tian0426/CL-HAR}.}. The library is highly modularized and easy to use, which opens up avenues for exploring novel contrastive models quickly in the future.

\end{abstract}



\maketitle

\section{Introduction}


The rapid proliferation of wearable devices such as mobile phones and fitness trackers have spurred on the increasing attention in sensing user activities and behavioural insights in numerous applications, e.g., healthcare, assisted living, wellness monitoring and smart building solutions~\cite{DBLP:journals/csur/ChenZYGYL21,DBLP:journals/imwut/ChangMISK20}. Wearable-based human activity recognition (HAR) seeks to accurately infer human activities based on streaming signals collected by wearable sensors, and it is playing an important role in ubiquitous and pervasive computing. In the past decade, deep learning models have outperformed traditional machine learning approaches in diverse application fields including HAR~\cite{DBLP:journals/prl/WangCHPH19,DBLP:conf/ijcai/QianPDM19,DBLP:journals/widm/RamamurthyR18}. Modern deep neural networks demonstrate outstanding performance especially when they are trained with a large number of labeled instances. However, collecting large volume of labeled activity data is a strong limiting factor since it is both expensive and time-consuming. Moreover, the label annotation process is prone to human bias and may result in ambiguous annotations. In order to alleviate the annotation issues, algorithms based on semi-supervised learning, weakly-supervised learning and transfer learning are proposed accordingly to alleviate the dependency of label annotations~\cite{DBLP:conf/aaai/QianPM19,DBLP:journals/ai/QianPM21,DBLP:journals/corr/abs-2006-03820}. While the above studies have sought to reduce the required label annotations under different learning paradigms, challenges persist in wearable-based tasks due to the difficulty of accurately annotating raw sensor signals which are not self-illustrative to human annotators compared with other modalities of data such as audio and images.

Recently, self-supervised learning establishes a new paradigm of learning feature representations without human annotations. Self-supervised learning designs a certain pretext task and generates corresponding intrinsic ground-truth labels for the pretext task automatically. According to the wearable sensors' characteristics, various pretext tasks can be constructed, such as masked reconstruction~\cite{DBLP:conf/iswc/HaresamudramBAG20}, motion forecasting~\cite{DBLP:journals/corr/abs-2010-13713}, and data transformation predictions~\cite{DBLP:journals/imwut/SaeedOL19}. Training models with pretext tasks can learn general-purpose latent representations, hence benefiting downstream tasks. Notably, we are witnessing an explosion of works based on contrastive learning for various applications, such as image classification, person re-identification, graph mining and natural language processing~\cite{DBLP:journals/corr/abs-2109-01116,DBLP:journals/access/Le-KhacHS20,DBLP:conf/nips/Sohn16}. Contrastive learning has emerged as a promising technique and achieves the state-of-the-art performance with instance discrimination as its pretext task, which even surpasses the supervised learning counterpart on downstream classification tasks in terms of accuracy~\cite{DBLP:conf/icml/ChenK0H20,DBLP:conf/nips/GrillSATRBDPGAP20}. Without human annotation, instance discrimination typically assigns the same label to samples augmented from the same instance, and assigns distinct labels to other instances.    


The implicit prerequisites of successful contrastive pretrained models, though, are the large-scale training samples as well as substantial computing resources such as TPUs~\cite{DBLP:conf/icml/ChenK0H20}. Therefore, it is not optimal to directly applying existing frameworks to the small-scale task of HAR. In addition, the unique data characteristics of wearables have not been fully investigated in contrastive models. Therefore, due to the above model scale and data characteristics differences, it is imperative to criticize and improve existing contrastive models for HAR tasks. Recently, empirical comparisons on data augmentation for HAR are conducted on specific models~\cite{DBLP:journals/corr/abs-2011-11542,DBLP:journals/corr/abs-2109-02054,DBLP:conf/icccn/LiuWLWYA21,DBLP:conf/ijcai/Wen0YSGWX21}. These studies indicate promising results of contrastive learning on HAR, but a deeper understanding of contributing ingredients in contrastive learning is still left mysterious.

To this end, we aim to shed light on what makes good contrastive learning on small-scale wearable-based HAR task through the lens of systematic algorithmic-level and task-level investigations. Specifically, for algorithmic-level analysis, we first decompose a general contrastive learning framework into several key ingredients, i.e., data augmentation transformations, backbone networks, construction of positive and negative pairs, projectors and predictors. For task-level investigations, we explore the challenges as well as opportunities to exiting models brought by wearables' data characteristics, such as cross-person generalization, robustness on wearing diversity and sliding window issues. Rigorous and controlled empirical evaluations are conducted on three benchmark datasets UCIHAR~\cite{DBLP:conf/iwaal/AnguitaGOPR12}, UniMiB SHAR (SHAR)~\cite{DBLP:journals/corr/MicucciMN16} and Heterogeneity Dataset for Human Activity Recognition (HHAR)~\cite{DBLP:conf/sensys/StisenBBPKDSJ15}. In the meantime, we release an easy-to-use open-source PyTorch library \texttt{CL-HAR}, featuring highly modularized contrastive learning components, standardized evaluation as well as systematic experiment management. We envision this work to provide valuable empirical evidence and insights of effective contrastive learning algorithms and to serve as a common testbed to foster future research.

The rest of the paper is organized as follows. The next section presents related work on HAR and contrastive learning algorithms. It is followed by discussions on a general paradigm of contrastive learning framework for HAR in Sec.~\ref{sec:general-cl}. Our detailed investigations on algorithmic level and task level are illustrated in Sec.~\ref{sec:analysis}. Sec.~\ref{sec:conclusion-discussion} recapitulates our observations and its significance.

\section{Related Work}\label{sec:related-work}

\textbf{Human Activity Recognition.}
In recent years, a large body of research for wearable-based human activity recognition is dedicated to learning discriminative features by leveraging various deep neural networks, including convolutional networks, residual networks, autoencoders and Transformers~\cite{DBLP:journals/prl/WangCHPH19,DBLP:journals/csur/ChenZYGYL21,DBLP:journals/sensors/MoralesR16,DBLP:conf/ijcai/QianPDM19,DBLP:journals/access/ShavitK21}. These models are accurate when the number of training instances is sufficiently large, yet such performance is not guaranteed when the labels are scarce. Semi-supervised learning methods seek to leverage unlabeled training instances which are typically easier to collect~\cite{DBLP:journals/sensors/OhALKK21,DBLP:conf/aaai/QianPM19,DBLP:conf/aaai/MaG19}. In addition, weakly-supervised learning approaches mitigate this issue by requiring a small amount of inaccurate or coarse labels~\cite{DBLP:journals/ai/QianPM21,DBLP:conf/smc/ShengH19}. Moreover, transfer learning approaches are proposed to transfer useful information from label-rich source domains to the label-sparse target domain~\cite{DBLP:journals/imwut/ChangMISK20,DBLP:conf/aaai/QianPM21,DBLP:journals/corr/abs-2006-03820,DBLP:journals/ijdsn/SaputriKL14}.

Recently, self-supervised learning, especially contrastive learning has demonstrated superior abilities in learning features in an unsupervised manner~\cite{DBLP:conf/iswc/HaresamudramBAG20,DBLP:journals/corr/abs-2010-13713,DBLP:journals/imwut/SaeedOL19}. Various forms of data transformation in contrastive learning have been investigated recently.~\cite{DBLP:journals/corr/abs-2011-11542} evaluated eight data augmentation techniques designed for wearable sensors in place of image augmentation operators in the SimCLR model.~\cite{DBLP:journals/corr/abs-2109-02054} investigated the efficacy of the sampling frequency of sensors and proposed a data augmentation technique based on re-sampling.~\cite{,DBLP:conf/icccn/LiuWLWYA21} applied both time-domain and frequency-domain augmentation techniques in SimCLR and the encoder was adapted from~\cite{DBLP:conf/www/YaoPJZSLLLWHS0A19} in order to be compatible with features in time and frequency domains. The Temporal and Contextual Contrasting (TS-TCC) method is proposed to incorporate the temporal characteristics of time series into contrastive learning~\cite{DBLP:conf/ijcai/Eldele0C000G21}. Instead of focusing only on data augmentation for a specific model, our work aims to provide rigorous empirical investigations on each component of contrastive learning. Besides, various types of contrastive learning models are simultaneously studied and compared.

\textbf{Contrastive Learning.}
Necessitated by the manual annotation bottleneck, contrastive learning creates different views of the same instance to form positive pairs in absence of labels. SimCLR obtains the positive sample from augmented views on the instance and contrast them against massive negatives~\cite{DBLP:conf/icml/ChenK0H20}. The key idea of instance discrimination is to treat each instance as a single category. Aside from generating positive samples from a single instance, other approaches assign samples from the same cluster or nearest neighbours as positives~\cite{DBLP:conf/iclr/0001ZXH21,DBLP:journals/corr/abs-2104-14548,DBLP:conf/nips/CaronMMGBJ20}. In order to eliminate the requirement of the massive number of negative samples, BYOL~\cite{DBLP:conf/nips/GrillSATRBDPGAP20} and SimSiam~\cite{DBLP:conf/cvpr/ChenH21} achieve competitive performance without any negative instance. The very recent literature provides taxonomy for existing approaches~\cite{DBLP:journals/access/Le-KhacHS20}.  

While there have been substantial gains from contrastive models, the reasons behind the gains are still unclear. This motivates research studies to demystify contrastive learning and unravel the hidden mechanisms behind its success. In~\cite{DBLP:conf/nips/Tian0PKSI20}, it is empirically shown that reducing the mutual information between the augmented samples while keeping task-relevant information intact can improve the overall representation learning.~\cite{DBLP:conf/nips/Purushwalkam020} found that even if existing approaches induce inconsistency during data augmentation, the learned models still demonstrate strong results for classification, which could be due to two reasons, i.e., a clean training dataset provides useful dataset bias for downstream tasks, and the capacity of representation function is low. In addition, the contrastive loss is analyzed to be closely related with two properties, i.e., i) the alignment of features from positive pairs and ii) the uniformity of the induced distribution of features on the hypersphere~\cite{DBLP:conf/icml/0001I20}.~\cite{DBLP:journals/corr/abs-2111-01124} studied the adversarial robustness of pretrained contrastive models. Meanwhile, contrastive learning is compatible with other learning frameworks. For instance,~\cite{DBLP:conf/nips/KhoslaTWSTIMLK20} proposed to extend the contrastive approach to the fully-supervised setting to leverage label information to complement the unsupervised nature of contrastive learning.~\cite{DBLP:conf/iclr/AgarwalMCB21} customized augmentation in reinforcement learning. The burgeoning studies on contrastive learning show extraordinary potency of learning from unlabelled data at scale, but they do not work well for small models, as pointed out by~\cite{DBLP:conf/iclr/FangWWZYL21}. Our analysis on wearable-based HAR task aims to fill the gap of contrastive models in small-scale tasks.

\section{Contrastive Learning in HAR}\label{sec:general-cl}





\subsection{Problem Formulation}

Given $N$ unlabeled samples $\{\mathbf{x}_{i}\}_{i=1}^{N}$, each sample $\mathbf{x}_{i} \in \mathbb{R}^{L\times D}$ denotes an activity which lasts for $L$ time-stamps and has $D$ dimensions. Due to the continuous stream of activity signals, the signals are usually segmented by fixed-size sliding window with window size of $L$ and step size of $\frac{L}{2}$ to form discrete samples. Our objective is to learn a neural-network-based encoder $f(\cdot;\phi)$ to produce latent features from these samples, i.e., $\mathbf{h}_i = f(\mathbf{x}_{i};\phi)$. The learned model can be subsequently utilized as a pretrained model for downstream tasks by fine-tuning a linear classifier on top of it. That is, after $f$ is trained, a downstream supervised classification task can be tackled by the classification network $\phi \circ f$, where the linear classifier $\phi(\cdot; \psi)$ is fine-tuned over the fixed backbone network $f$.


\subsection{A General Contrastive Learning Framework}
As suggested by its name, the contrastive model is trained by learning to contrast between different instances with the instance discrimination task. It is achieved by treating each individual instance as an independent class. Samples generated from a specific instance by data augmentation are treated as positives, while rest instances are treated as negative samples. Positives are pulled closer to each other while negatives are pulled away from the instance. 

A data augmentation transformation function $T(\cdot):\mathcal{X}\rightarrow \mathcal{X}$ indicating a single random transformation or a compositional transformation is applied to obtain augmented views $\{\mathbf{x}_{i}^{1}, \mathbf{x}_{i}^{2}\}$ for each $\mathbf{x}_{i}$. We then obtain latent representations $\{\mathbf{h}_{i}^{1}, \mathbf{h}_{i}^{2},...\}$ for these views using the backbone encoder network $f$. The parameters of networks are omitted for simplicity hereinafter. A projection head $g(\cdot)$ is usually utilized to map $\mathbf{h}$ into a lower-dimensional embedding space $\mathbf{z} = g(\mathbf{h})$. In the setting of unsupervised learning, the ground truth class label for $\mathbf{x}_{i}$ is unavailable. The objective of contrastive learning is to map input instances $\{\mathbf{x}_{i}\}_{i=1}^{N}$ to compact feature representations $\{\mathbf{z}_{i}\}_{i=1}^{N}$ where similar instances are closer while dissimilar instances are separated. To be concrete, the InfoNCE loss for the positive pair $(\mathbf{x}_{i}^{1}, \mathbf{x}_{i}^{2})$ is defined as
\begin{equation}
\label{eq:contrastive-loss}
    \ell_{i} = - \log \frac{\exp(\text{sim}(z_i^{1}, z_i^2)/\tau)}{\exp(\text{sim}(z_i^{1}, z_i^2)/\tau)+\sum_{k = 1}^{N}\mathbbm{1}_{[k \neq i]} \exp(\text{sim}(z_i^{1}, z_k)/\tau)}
\end{equation}
where $\mathbbm{1}_{[k \neq i]} \in \{0, 1\}$ is an indicator function, sim$(\cdot,\cdot)$ denotes a similarity metric, $(z_i^{1}, z_k)$ is any negative pair and $\tau$ denotes the temperature parameter. Such loss function in Eq.~\eqref{eq:contrastive-loss} is computed across all positive pairs, i.e., $\mathcal{L}_{\text{total}} = \sum_{i=1}^{N} \ell_{i}$.



\subsection{Key Components and Discussions}
\subsubsection{Data Augmentation Transformations}

The choice of transformation functions $T$ controls the properties of the learned invariance in representations. Most existing contrastive models are inspired by experiences in computer vision domains, which implicitly contain strong inductive bias such as cropping-invariance on image data. Nevertheless, for activity data, such invariance is sub-optimal since cropping a subset of time-series data may inevitably lose crucial temporal cues. Taking this cue, we seek to study data transformation functions that are tailored for time-series instances, as listed in Table~\ref{table:aug-funcs}. Specifically, two categories of data transformations are investigated, i.e., time-domain and frequency-domain functions. The time-domain transformation functions stem from wearable-based time series, and are widely adopted in self-supervised approaches in HAR~\cite{DBLP:conf/icmi/UmPPELHFK17,DBLP:journals/corr/abs-2011-11542}. It consists of both single and compositional functions. The augmentation in the frequency domain is inspired by the fact that wearable devices measure physical phenomena, where collected data are fundamentally a function of signal frequencies~\cite{DBLP:conf/www/YaoPJZSLLLWHS0A19}. In this paper, we investigate four frequency-based transformations adapted from~\cite{DBLP:conf/icccn/LiuWLWYA21}. The main difference between our analysis and ~\cite{DBLP:conf/icccn/LiuWLWYA21} is that, ~\cite{DBLP:conf/icccn/LiuWLWYA21} keeps features in frequency domain, while ours transform the augmented instances back to the time domain to feed into backbone networks, which circumvents the modification of existing backbone networks. To be specific, each instance is transformed to the frequency domain by Fast Fourier Transform (FFT)~\cite{bracewell1986fourier}. The frequency response at frequency $\omega_{k}=2\pi k/L$ is defined as
\begin{equation}
F(\omega_{k})=\frac{1}{L} \sum_{t=1}^{L} x_{t} e^{-j \omega_{k} t}=A(\omega_{k}) e^{j \theta(\omega_{k})}, k \in \{1, 2 \ldots, L\},
\end{equation}
where $A(\omega_{k})$ and $\theta(\omega_{k})$ are the amplitude spectrum and phase spectrum, respectively. The frequency-based augmentation is applied on $F(\omega_{k})$. 
Then the data are transformed back to the time domain by inverse Fast Fourier Transform (iFFT)
\begin{equation}
\hat{x_{t}}=\frac{1}{L} \sum_{k=1}^{L} F(\omega_{k}) e^{j \omega_{k} t}, t\in \{1,...,L\}.
\end{equation}

\begin{table*}
\caption{Details of Data Augmentation Functions in the Experiments. }
\label{table:aug-funcs}

\begin{tabular}{p{0.08\textwidth} 
p{0.1\textwidth}  p{0.7\textwidth}}
\toprule
Domain   & Augmentation & Implementation Details \\
  \midrule 
\multirow{2}{*}{Time} & noise  & add a randomly generalized noise signal with a mean of 0 and standard deviation of 0.8   \\ 
  & scale         & amplify each channel by a randomly generalized distortion with a mean of 2 and standard deviation of 1.1   \\ 
  & shuffle     &  randomly permute the channels of the sample \\     
   & negate         &  multiply the value of the signal by a factor of -1       \\      
 & permute       &  randomly split each signal into no more than 5 segments in the time scale, then permute the segments and combine them into original shape     \\ 
  & resample         &  up-sample the signal in time axis to 3 times its original time steps by linear interpolation and randomly down-sample to its initial dimensions     \\ 
  & rotation      &  rotate the 3-axial (x, y and z) readings of each sensor by a random degree, which follows a uniform distribution between $-\pi$ and $\pi$, around a random axis in the 3D space   \\ 
         
 & t\_flip  &   reverse the signal in time dimension       \\ 
 & t\_warp     &   stretch and warp each signal in the temporal dimension with an arbitrary cubic spline       \\ 
 & perm\_jit          &  apply permutation and noise      \\          
  & jit\_scal     &   apply noise and scaling      \\ \hline
\multirow{3}{*}{Frequency} & hfc   &   split the low and high frequency components and reserve high frequency components  \\ 
     & lfc         &  split the low and high frequency components and reserve low frequency components  \\ 
    & p\_shift    &   shift the phase values of the frequency response with a randomly generalized number from the uniform distribution between $-\pi$ and $\pi$      \\ 
     & ap\_p    &  perturb the amplitude and phase values of a randomly selected segment of the frequency response of the sample. The selected segment is of half length of the frequency domain data. The perturbation on amplitude is a Gaussian noise with a mean of 0 and standard deviation of 0.8. The perturbation on phase follows a randomly generalized uniform distribution between $-\pi$ and $\pi$   \\ 
 & ap\_f & apply the amplitude and phase perturbation to the whole sequence of frequency response with the same perturbation ranges as ap\_p's\\
\bottomrule
\end{tabular}
\end{table*}

\subsubsection{Backbone Networks}

The backbone networks are usually fixed to be ResNet in existing works, and hence are less investigated than data augmentation in contrastive learning. We reckon that a proper backbone network is critical to learning representations in HAR and we implement and compare six commonly utilized neural networks whose representation capabilities conform to wearables' data characteristics, i.e., DeepConvLSTM, LSTM, CNN, AE, CAE and Transformer. Details are presented in Appendix~\ref{appendix:implementation-details}.

\subsubsection{Construction of Positive and Negative Pairs}

How to construct the positive pairs in Eq.~\eqref{eq:contrastive-loss} is crucial to contrastive learning, since the models are encouraged to be invariant to transformations among positive pairs. Current contrastive learning approaches treat different views of the same instance as positives, and such views are usually generated by applying predefined transformations on the instance. For existing models SimCLR, BYOL, SimSiam and TS-TCC, $(\mathbf{x}_{i}^{1}, \mathbf{x}_{i}^{2})$ are considered as a positive pair, and we denote such scenario as \textit{2augs}. We also study the alternatives denoted as \textit{1aug}, i.e., $(\mathbf{x}_{i}^{1}, \mathbf{x}_{i})$ or $(\mathbf{x}_{i}^{2}, \mathbf{x}_{i})$, since $\{\mathbf{x}_{i}^{1}, \mathbf{x}_{i}^{2}, \mathbf{x}_{i}\}$ are expected to convey the same semantic content.

While it is convenient to construct such positive pairs from the same instance, it inevitably ignores the inter-sample relations. Hence, it is also plausible to construct positive pairs from distinct samples in the dataset. For instance, the Nearest-Neighbour Contrastive Learning of visual Representations (NNCLR) method finds the nearest neighbour of the sample from a memory bank to be the positive pairs, i.e., 
\begin{equation}
\ell_{i}^{\text{NNCLR}} = - \log \frac{\exp(\text{sim}(z_i,\text{NN}(z_i, Q))/\tau)}{\sum_{k = 1}^{N}\exp(\text{sim}(\text{NN}(z_i, Q), z_k)/\tau)},
\end{equation}
where $\text{NN}(z_i, Q)$ is the nearest neighbour of $z_i$ based on distance metric $Q$.


The SimCLR, NNCLR and TS-TCC require negative pairs for training, which can be obtained within mini-batches or memory banks. BYOL learns a high-quality representation without negative samples. To achieve so, BYOL trains an online network to predict the target network representation of the same instance under different views and using an additional predictor network on top of the online encoder to prevent the model collapse. SimSiam further validates the design without negatives when the momentum encoders are removed and the size of batch size is reduced. We seek to compare these models, and investigate the influence of pair sizes.

\section{Empirical Investigations}\label{sec:analysis}
\subsection{Algorithmic-Level Investigations}~\label{sec:alg-level-analysis}

Our experiments are implemented with PyTorch~\cite{DBLP:conf/nips/PaszkeGMLBCKLGA19} and we report accuracy performance metrics. Other metrics are available in \texttt{CL-HAR}. Implementation details can be found in Appendix~\ref{appendix:implementation-details}.

\subsubsection{Effects of Data Augmentation}\label{subsec:augmentation}
\begin{figure*}[!h]
    \subfigure[performance on UCIHAR dataset]{\label{fig:2-augs-heatmaps-b}\includegraphics[width=0.45\linewidth]{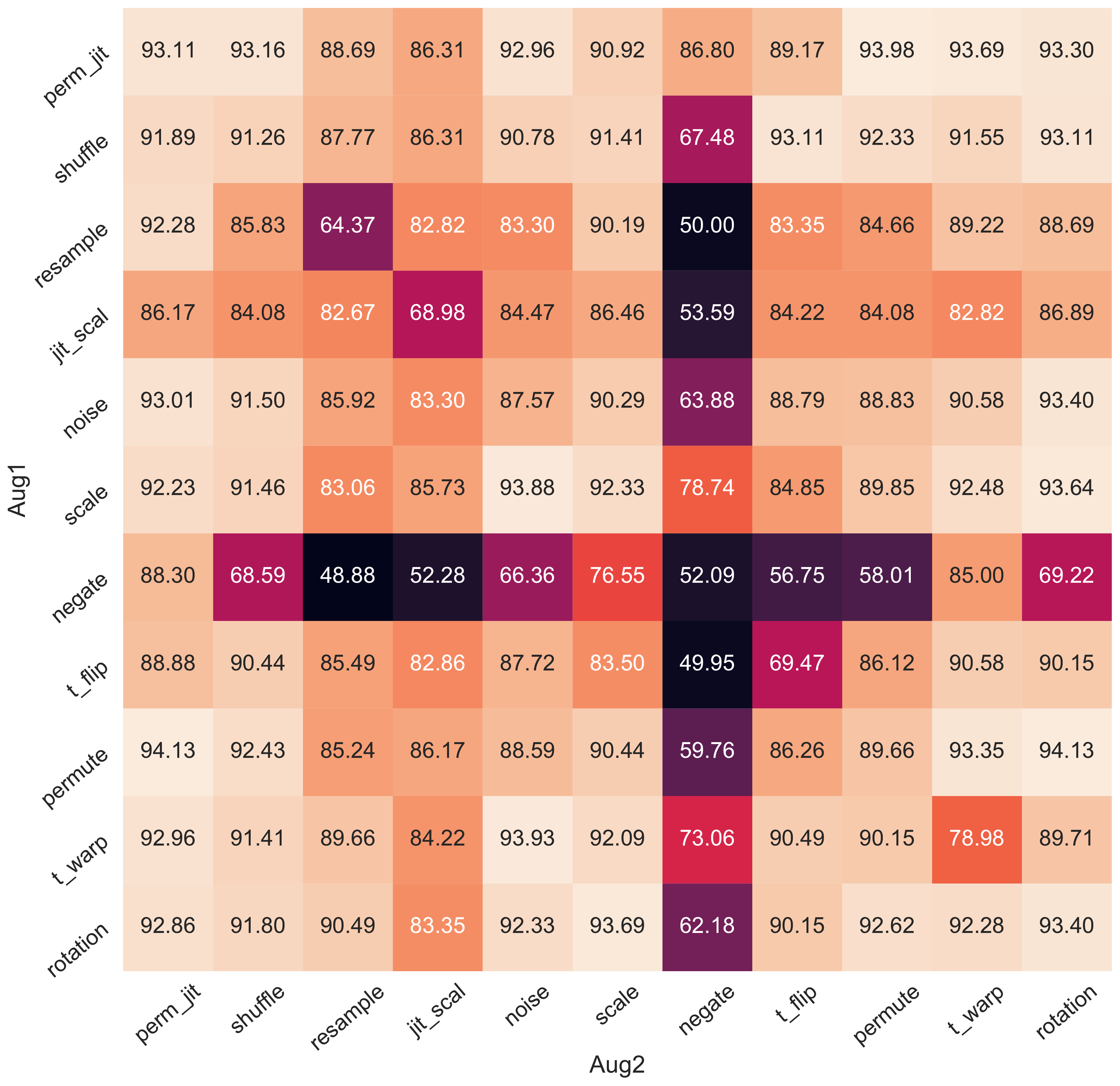}}
    \hspace{0.01cm}
    \subfigure[performance on SHAR dataset]{\label{fig:2-augs-heatmaps-b}\includegraphics[width=0.45\textwidth]{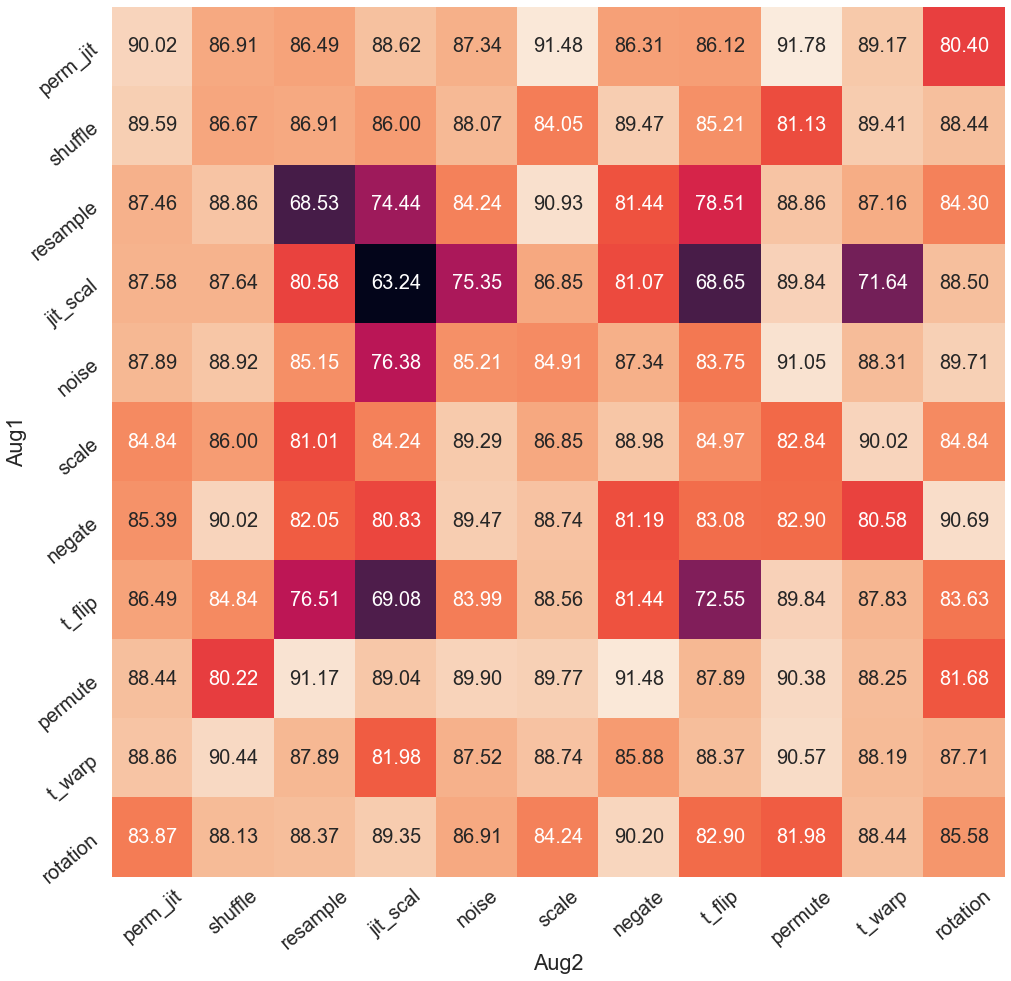}}
  
  \caption{Visualization of various combinations of augmentation functions on SimCLR model on (a) UCIHAR and (b) SHAR datasets.}\label{fig:2-augs-heatmaps}
\end{figure*}

It is observed empirically that different transformation functions of data augmentation can affect the performance of contrastive learning approaches~\cite{DBLP:journals/corr/abs-2011-11542}. We compare time-domain augmentation under \textit{1aug}, and the results on UCIHAR and SHAR datasets are listed in Table~\ref{table:data-augmentation-ucihar} and Table~\ref{table:data-augmentation-shar} respectively. It is observed that there is no unique augmentation transformation that consistently performs better than others in all models. When two positives are both generated by augmentation in \textit{2augs}, as shown in Fig.~\ref{fig:2-augs-heatmaps}, the variations of performance are enlarged. It implies that some augmentation functions fail to preserve the semantic meaning of the original instance and cause the distortion of semantics. The results on frequency-domain augmentation are listed in Table~\ref{table:data-augmentation_freq-ucihar} and Table~\ref{table:data-augmentation_freq-shar}, and variations on results exist as well. The caveat is that invariance induced by augmentation should not be taken for granted, and how to seek a proper augmentation transformation automatically is still an open problem.


\begin{table*}[!htbp]
  \caption{Comparisons on data augmentation functions in contrastive learning approaches in UCIHAR dataset.}
  \label{table:data-augmentation-ucihar}
  \begin{tabular}{l l l l l l l l l l l l l}
    \toprule
Models  & noise & scale & negate & perm & shuffle & t\_flipped & t\_warp & resample & rotation & perm\_jit & jit\_scal & mean $\pm$ std\\
    \midrule
BYOL    & \textbf{94.17}  & 88.16   & 84.95   & 90.73       & 83.20   & 92.18         & 89.03      & 91.7     & 88.2     & 89.90     & 89.08   & 89.21 $\pm$ 3.14  \\
SimSiam & 83.93 & 82.91 & 80.97 & 85.63 & 80.78 & \textbf{89.51} & 83.06 & 88.25 & 87.57 & 81.41 & 81.02  & 84.09 $\pm$ 3.18   \\
SimCLR  & \textbf{93.64}  & 90.0    & 81.94   & 88.25       & 74.9    & 90.15         & 88.98      & 91.84    & 54.08    & 89.51     & 93.54    & 85.17 $\pm$ 11.66 \\
NNCLR   & 90.53  & 86.31   & 85.24   & 85.39       & 60.92   & 85.24         & 84.95      & \textbf{95.10}    & 51.31    & 85.29     & 89.56   & 81.80 $\pm$ 13.26  \\
TS-TCC  & 93.40  & 89.17   & 90.34   & 90.29       & 92.14   & 89.08         & 90.58      & 91.65    & 89.61    & \textbf{93.79}     & 90.97  & 91.00 $\pm$ 1.59   \\
  \bottomrule
\end{tabular}
\end{table*}

\begin{table*}
  \caption{Comparisons on data augmentation functions in contrastive learning approaches in SHAR dataset.}
  \label{table:data-augmentation-shar}
  \begin{tabular}{l l l l l l l l l l l l l}
    \toprule
Models  & noise & scale & negate & perm & shuffle & t\_flipped & t\_warp & resample & rotation & perm\_jit & jit\_scal & mean $\pm$ std \\
    \midrule
BYOL    & 89.41 & 88.86 & 89.59 & 89.29 & 89.90 & 87.46 & 89.96 & 88.86 & 84.84 & \textbf{91.17} & 87.71 & 88.82 $\pm$ 1.68\\
SimSiam & 89.17 & 89.17 & 89.59 & 87.52 & 88.62 & 89.41 & \textbf{90.69} & 87.34 & 87.22 & 86.37 & 88.68 & 88.53 $\pm$ 1.27\\
SimCLR  & 87.10 & 84.60 & 82.90 & 84.72 & 77.24 & 70.66 & 88.13 & 88.56 & 81.07 & \textbf{88.62} & 84.42 & 83.46 $\pm$ 5.48 \\
NNCLR   & 85.88 & 84.18 & 88.62 & 87.83 & 87.16 & \textbf{90.20} & 84.18 & 89.41 & 83.93 & 89.35 & 84.48 & 86.84 $\pm$ 2.40\\
TS-TCC  & 79.37  & 70.72   & 79.00   & 83.51       & 79.12   & 82.47 & 79.85  & 80.10    & 78.27    & \textbf{83.81}     & 68.05   & 78.57 $\pm$ 4.95  \\
  \bottomrule
\end{tabular}
\end{table*}

\begin{table}
  \caption{Comparisons on frequency-domain augmentation functions in contrastive learning in UCIHAR dataset.}
  \label{table:data-augmentation_freq-ucihar}
\begin{tabular}{lllllll}
\toprule
Models   & hfc   & lfc   & p\_shift & ap\_p & ap\_f & mean $\pm$ std \\ 
        \midrule
BYOL    & 86.70 & 87.04 & 91.50 & 91.55 & \textbf{91.70} & 89.70 $\pm$ 2.59 \\
SimSiam & 86.36 & 83.01 & 87.71 & 90.24 & \textbf{91.26} & 87.72 $\pm$ 3.28  \\
SimCLR  & 83.54 & 86.21 & \textbf{91.89} & 89.42 & 88.88 & 87.99 $\pm$ 3.20 \\
NNCLR   & \textbf{82.86} & 81.07 & 81.26 & 80.92 & 71.50 & 79.52 $\pm$ 4.55 \\
TS-TCC  & \textbf{94.37} & 92.14 & 92.33 & 91.07  & 92.37 & 92.46 $\pm$ 1.19  \\                 

\bottomrule
\end{tabular}
\end{table}

\begin{table}
  \caption{Comparisons on frequency-domain augmentation functions in contrastive learning in SHAR dataset.}
  \label{table:data-augmentation_freq-shar}
\begin{tabular}{lllllll}
\toprule
Models    & hfc   & lfc   & p\_shift & ap\_p & ap\_f & mean $\pm$ std \\ 
        \midrule
BYOL    & 79.06 & \textbf{89.65} & 72.98    & 77.6    & 71.82   & 78.22$\pm$  7.07    \\
SimSiam & 85.09 & \textbf{90.14} & 85.58    & 77.97  & 88.92   & 85.54$\pm$  4.75   \\
SimCLR  & 69.99 & 65.12 & \textbf{85.09}    & 64.09    & 76.57   & 72.17$\pm$ 8.75  \\
NNCLR   & 74.5  & \textbf{90.57} & 83.87    & 75.59    & 89.35    & 82.77 $\pm$  7.50   \\
TS-TCC  & 69.87 & 81.8  & \textbf{82.96}  & 74.25    & 82.90    & 78.35 $\pm$ 5.97   \\
\bottomrule
\end{tabular}
\end{table}

\subsubsection{Effects of Backbone Networks}\label{subsec:encoder}

\begin{figure*}[!h]
  \centering
  \includegraphics[width=0.9\linewidth]{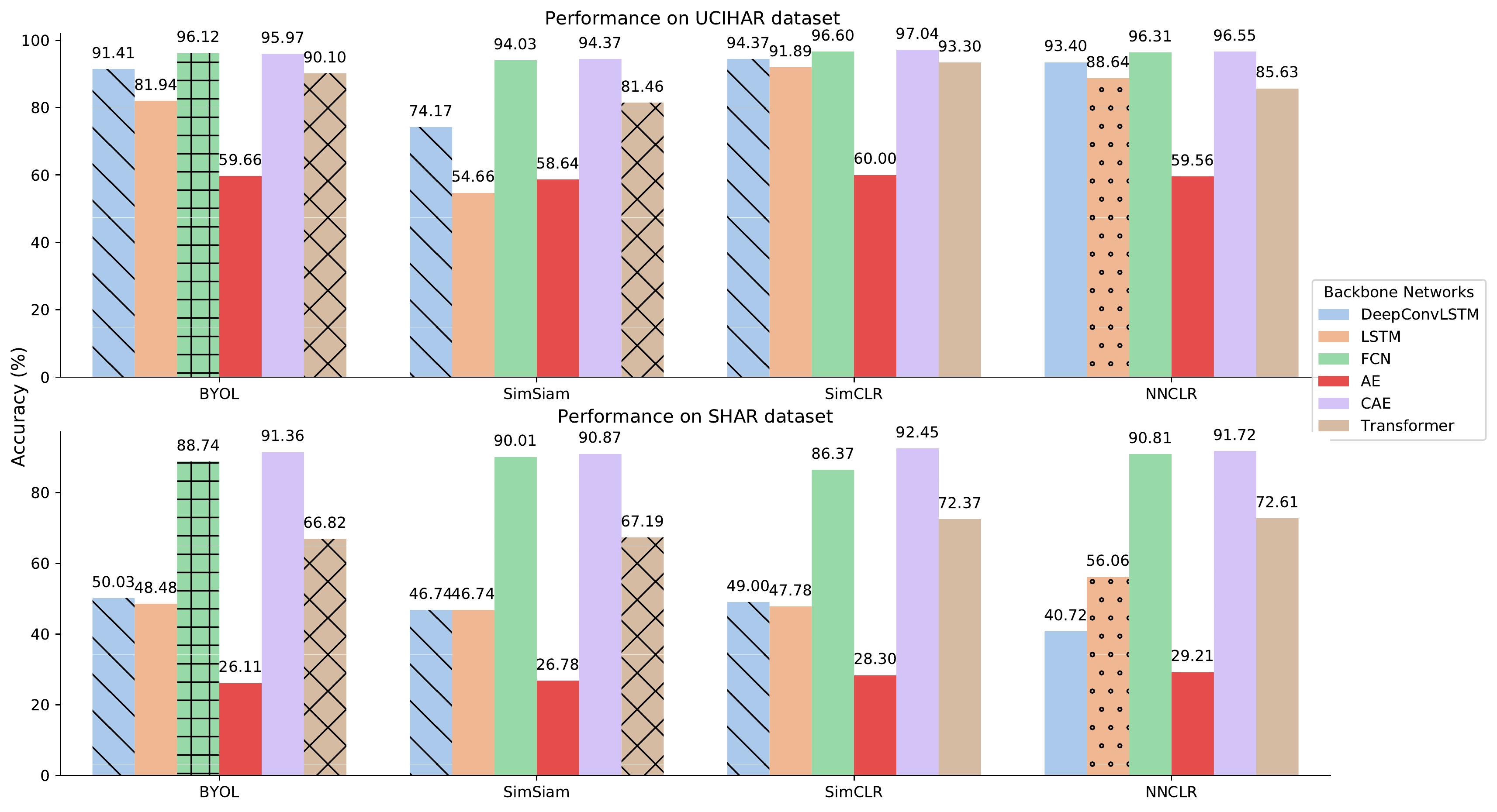}
  \caption{Visualization of performance on four contrastive models with six different backbone networks.}
  \label{fig:backbone-results}
\end{figure*}

\begin{figure*}[!h]
    \subfigure[Effects of number of convolutional layers]{\label{fig:x}\includegraphics[width=0.4\linewidth]{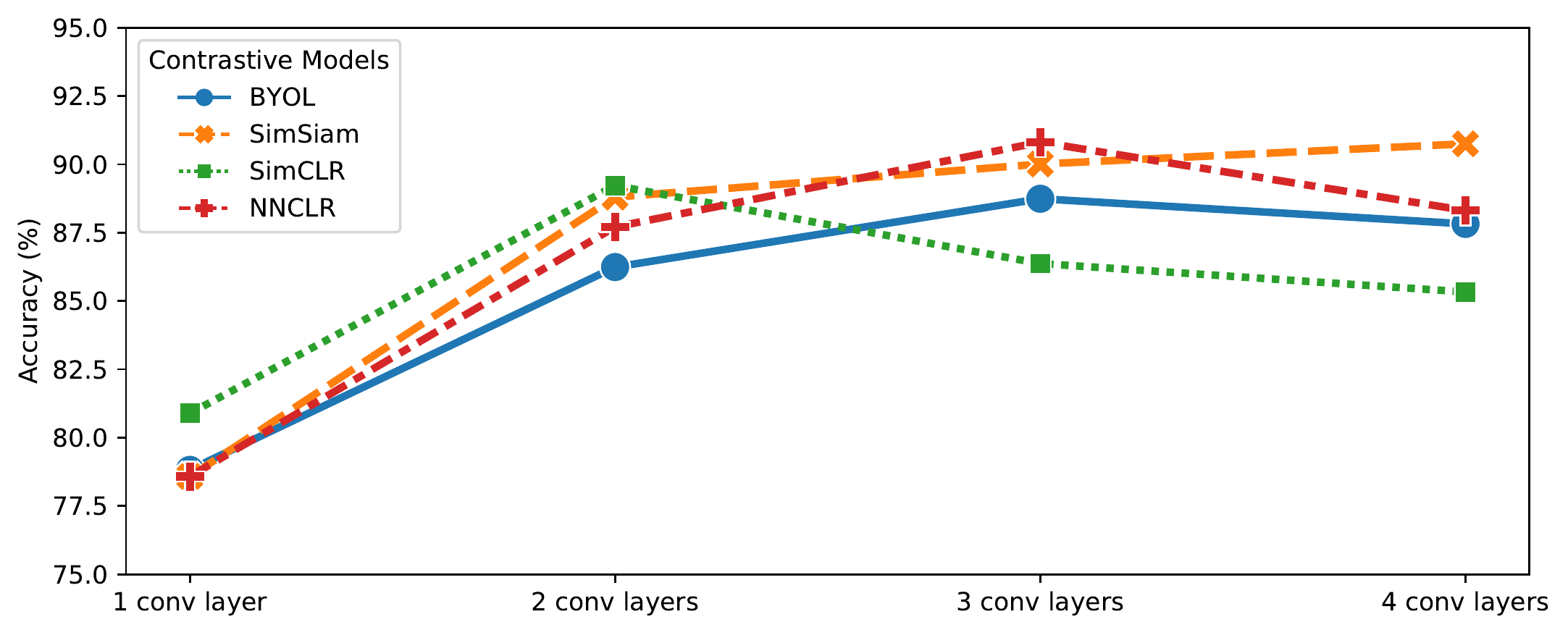}}
    \hspace{0.02cm}
    \subfigure[Variants of FCN.]{\label{fig:xx}\includegraphics[width=0.4\textwidth]{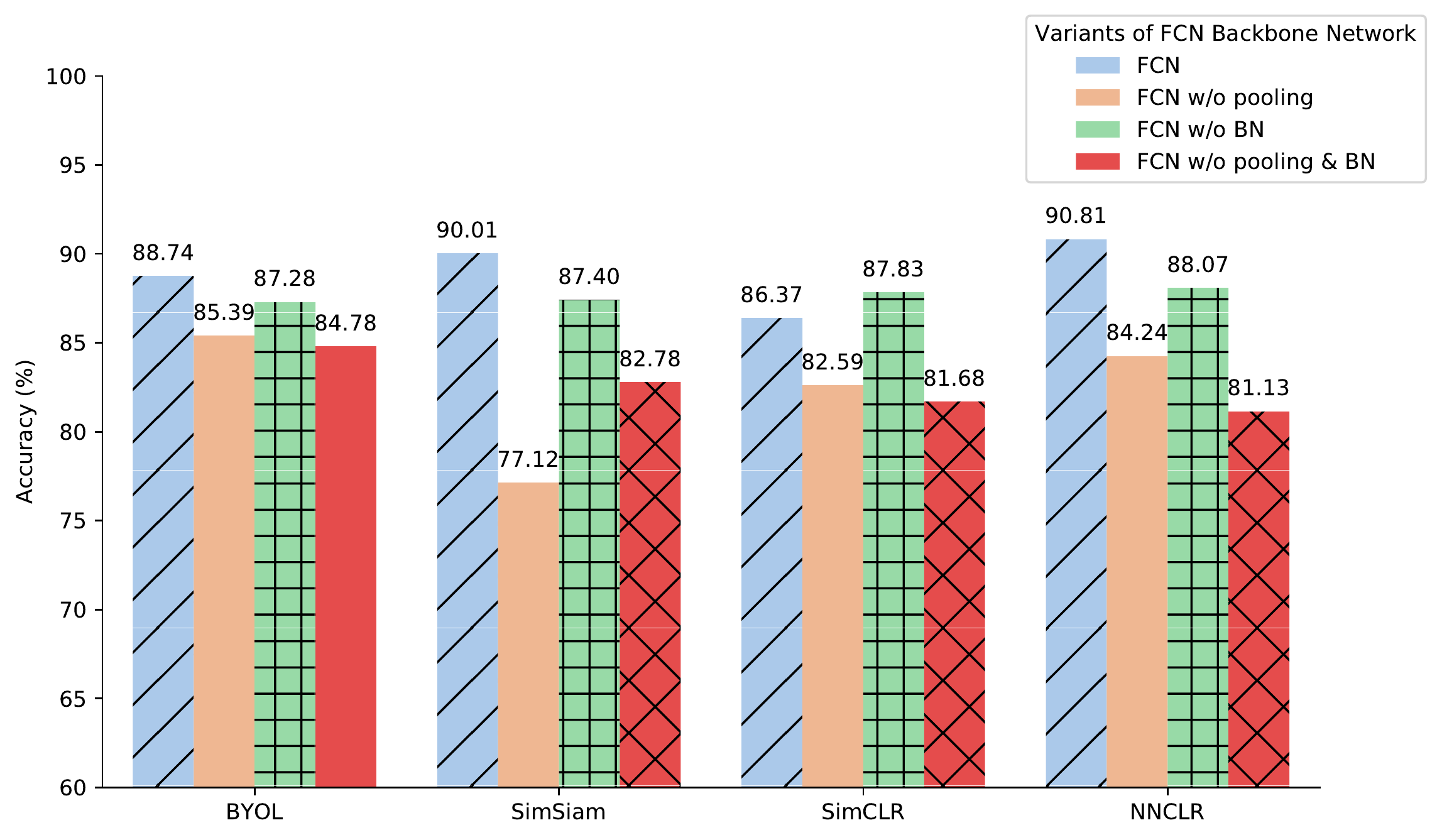}}
  \caption{Visualization of ablation studies on FCN backbone network in SHAR dataset. (a) effects of different number of convolutional layers. (b) effects of batch normalization and pooling layers.}\label{fig:backbone-fcn-ablation}
\end{figure*}

To investigate the effects of the backbone networks, we ablate six commonly adopted neural networks for each contrastive model. The results are depicted in Fig.~\ref{fig:backbone-results}. On UCIHAR dataset, convolution-based networks (FCN and CAE) are slightly better than LSTM-based networks (DeepConvLSTM and LSTM). While on SHAR dataset, convolution-based networks are significantly better than LSTM-based networks. AE performs the worst on both datasets, which may due to its limited network capacity of linear layers. Transformer performs worse than convolutions, while achieves better results than LSTM- and linear-based models. Therefore, convolutional neural networks are more robust than other networks for HAR. Meanwhile, the reconstruction loss in CAE is complementary to contrastive loss, as CAE is slightly better than FCN in most cases. In addition, we study the effects of FCN's model capacity w.r.t. the number of convolutional layers, as well as the effects of pooling and batch normalization (BN) layers. The results are illustrated in Fig.~\ref{fig:backbone-fcn-ablation}. Our ablations suggest that i) proper network capacities are model-specific, and in general, larger capacities are beneficial to models, ii) pooling layer is beneficial to contrastive models, and iii) BN layer helps the model in most cases. 

\subsubsection{Effects of Negative Pairs}
\begin{figure}[!h]
    \subfigure[batch size of SimCLR]{\label{fig:negative-pairs-bs}\includegraphics[width=0.43\linewidth]{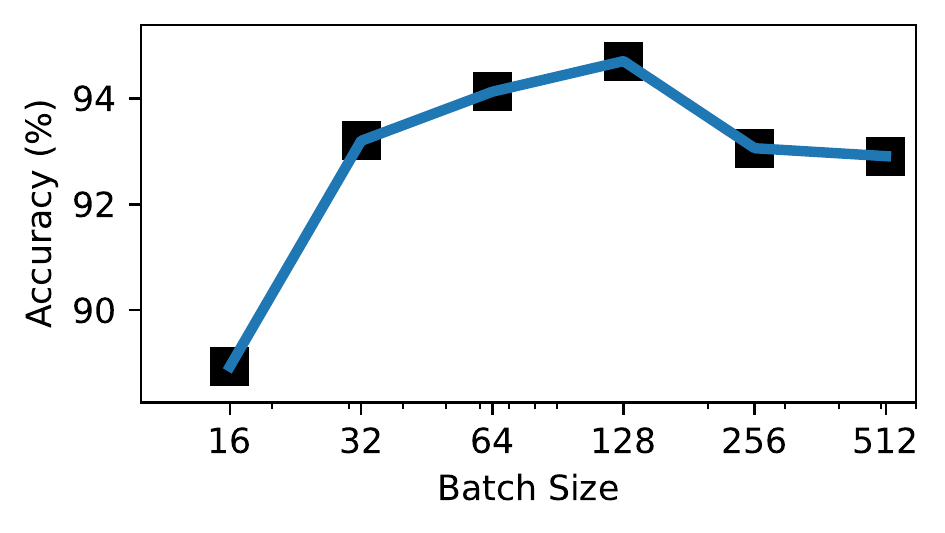}}
    \hspace{0.01cm}
    \subfigure[support set in NNCLR]{\label{fig:negative-pairs-memory}\includegraphics[width=0.21\textwidth]{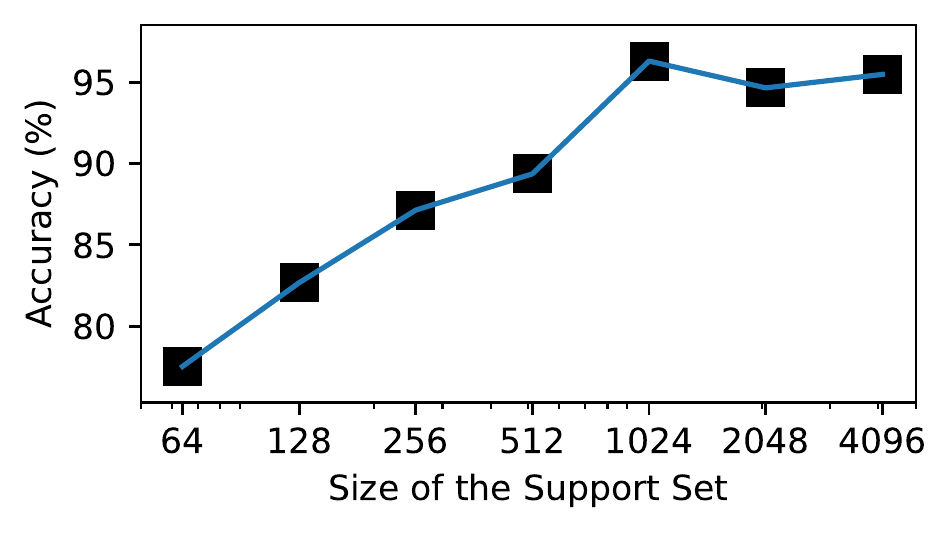}}
  \caption{Comparisons of performance on UCIHAR dataset}\label{fig:negative-pairs}
\end{figure}

SimCLR treats other instances within a batch as negatives, hence we alter the batch sizes to validate the effects of negative pairs. As shown in Fig.~\ref{fig:negative-pairs-bs}, increasing the number of negatives from 16 to 128 can greatly improve the performance. Further enlarging the batch size from 128 instead hurts the performance. This can be caused by the inherent inconsistency of instance discrimination, where all instances within a batch are considered as different classes, despite the fact that many of them actually share the same class. For NNCLR, such performance degradation is less obvious, since enlarging the size of support set increases the possibility of finding a better nearest neighbour, hence improving the positive pairs.


\subsubsection{Positive Pairs from Distinct Samples}
On UCIHAR, NNCLR can achieve the best performance of 95.1\% as shown in Table~\ref{table:data-augmentation-ucihar}. With a larger support set, its performance can achieve 96\% in Fig.~\ref{fig:negative-pairs-memory}. This favorably advocates that semantically consistent positive pairs are crucial in contrastive learning. On SHAR, however, such advantage is less obvious. The possible reason is that larger data discrepancies in SHAR increases the difficulty of defining a good nearest neighbour.


\subsubsection{Effects of Projector and Predictor}
\begin{figure}[!h]
    \subfigure[batch size of SimCLR]{\label{fig:projector}\includegraphics[width=0.43\linewidth]{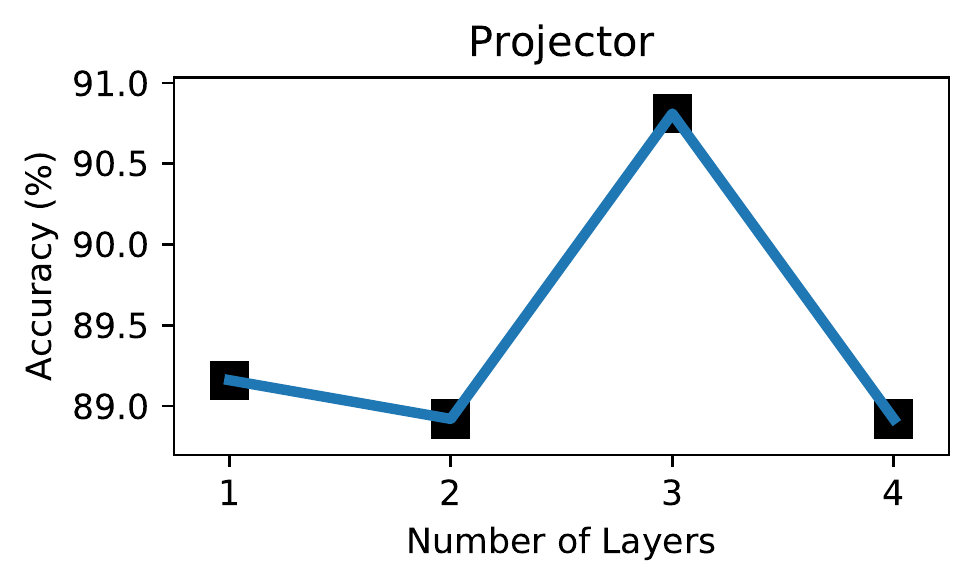}}
    \hspace{0.01cm}
    \subfigure[support set in NNCLR]{\label{fig:predictor}\includegraphics[width=0.21\textwidth]{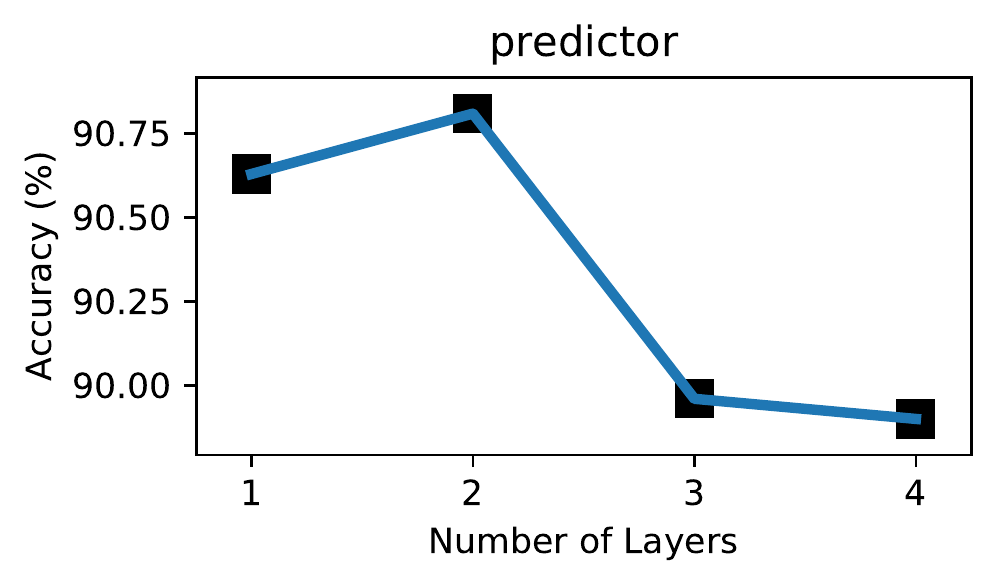}}
  \caption{Effects of capacities of the projector and predictor of NNCLR on SHAR with the backbone network being FCN.}\label{fig:effects-projector-predictor}
\end{figure}

As illustrated in Fig.~\ref{fig:effects-projector-predictor}, with the number of layers ranging from 1 to 4 in both the projector and predictor, the accuracy is ranged from 89\% to 91\%. Therefore, the two components are less critical in contrastive models.

\subsection{Task-Level Investigations}~\label{sec:task-level-analysis}
We conduct experiments from task-relevant perspective which is closely tied to the data characteristics of wearables.

\begin{table}[!htbp]
  \caption{Performance of contrastive models on UCIHAR under cross-person setting. The source domain indices are $\{0,..., 29 \setminus \text{target domain index} \}$.} 
  \label{table:cross-person-ucihar-large}
\begin{tabular}{llllll}
\hline
\toprule
Target  & 0   & 1    & 2    & 3    & 4  \\ \hline
BYOL          & \textbf{100.00} & 92.38     & 99.12         & 96.21         & \textbf{93.38}     \\
SimSiam       & \textbf{100.00} & 91.72     & 98.83         & 94.32         & 92.72     \\
SimCLR        & \textbf{100.00} & \textbf{95.70}     & 97.95         & \textbf{96.53}         & 92.05     \\
NNCLR         & \textbf{100.00} & 92.43     & 98.53         & 92.43         & 92.38     \\
TS-TCC        & 99.42  & 88.74     & \textbf{99.41}         & 93.06         & 84.44     \\ 
\bottomrule
\end{tabular}
\end{table}

\begin{table}[!htbp]
  \caption{Performance of contrastive models on SHAR under cross-person setting. The source domain indices are $\{1, 2, 3, 5, 6, 9, 11, 13-17, 19-25, 29 \setminus \text{target domain index} \}$.}
  \label{table:cross-person-shar-large}
\begin{tabular}{lllll}
\hline
\toprule
Target Domain & 1    & 2     & 3    & 5      \\ \hline
BYOL  & \textbf{69.53}  & 67.07   & 76.32   & \textbf{74.16}   \\
SimSiam  & 65.36  & 67.41  & 75.33  & 69.46     \\
SimCLR   & 66.67 & \textbf{73.58}   & 74.67  & \textbf{74.16}    \\
NNCLR  & 66.67   & 71.87   & \textbf{81.25}   & 73.49 \\
TS-TCC    & 60.16  & 62.44  & 75.66   & 69.80   \\ 
\bottomrule
\end{tabular}
\end{table}

\begin{figure*}[!h]
    \subfigure[Window length:50]{\label{fig:slidwin50}\includegraphics[width=0.24\linewidth]{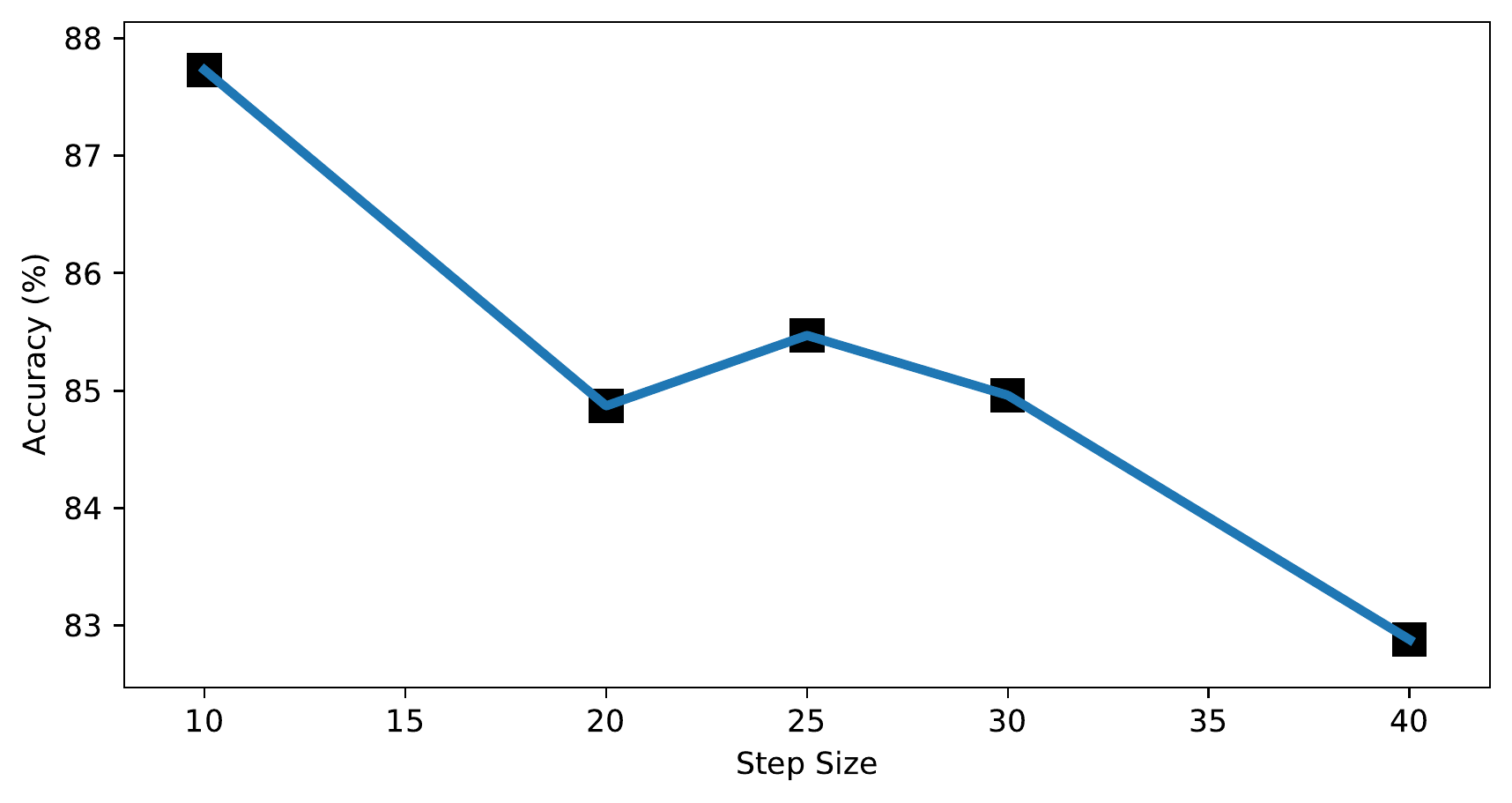}}
    \hspace{0.01cm}
    \subfigure[Window length:100]{\label{fig:slidwin100}\includegraphics[width=0.24\textwidth]{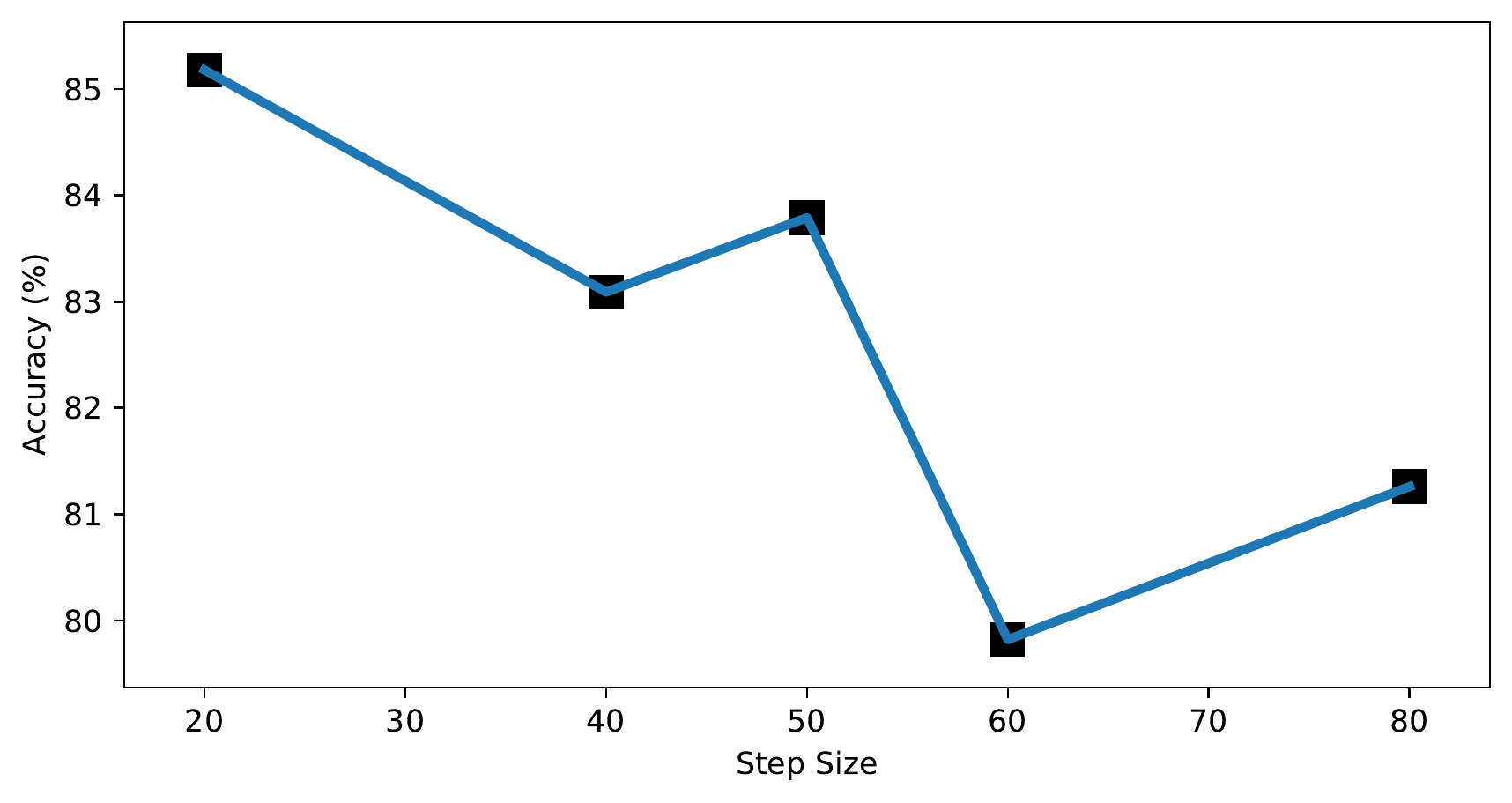}}
    \hspace{0.01cm}
  \subfigure[Window length:200]{\label{fig:slidwin200}\includegraphics[width=0.24\textwidth]{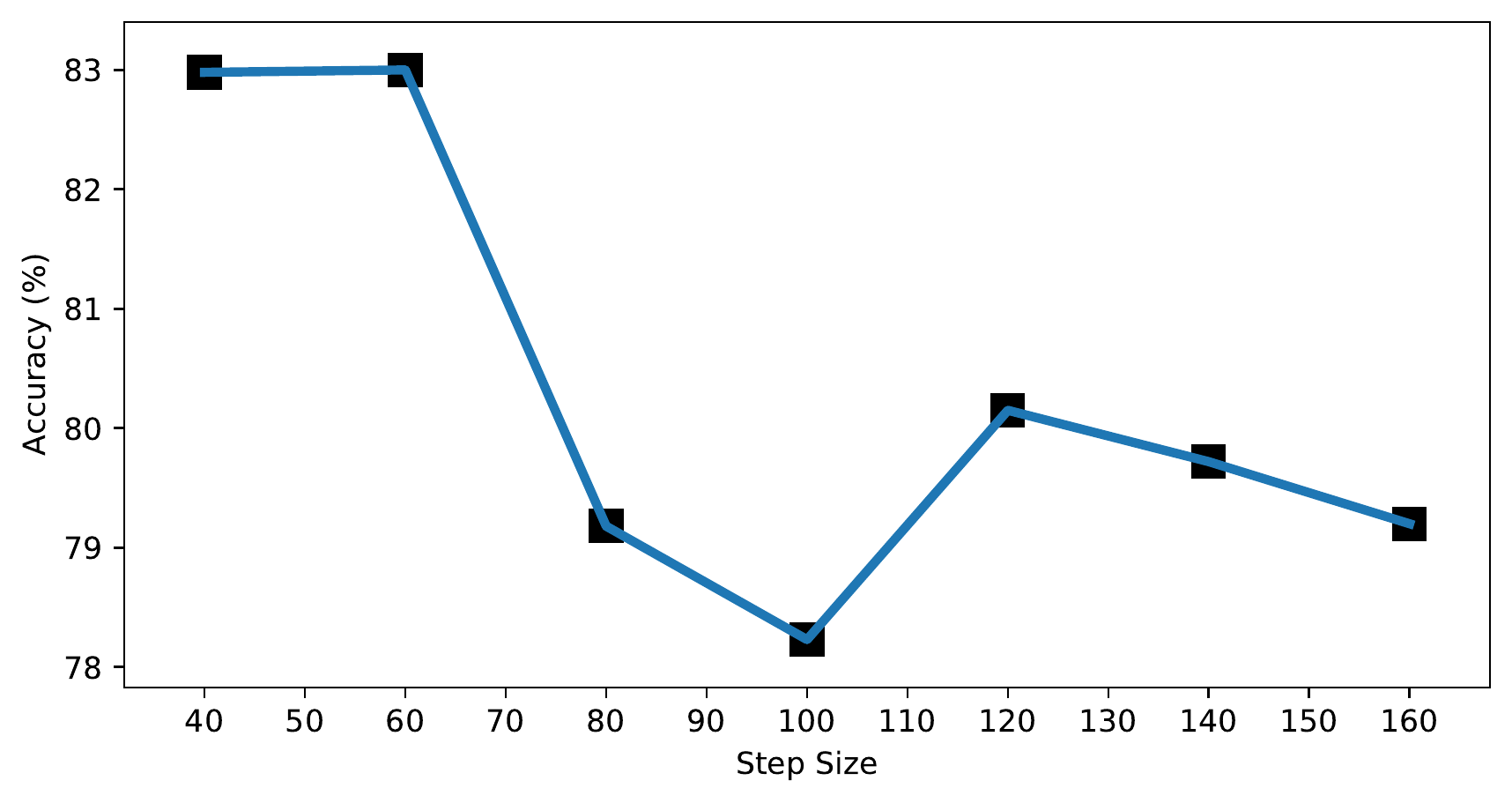}}
    \hspace{0.01cm}
  \subfigure[Window length:400]{\label{fig:slidwin400}\includegraphics[width=0.24\textwidth]{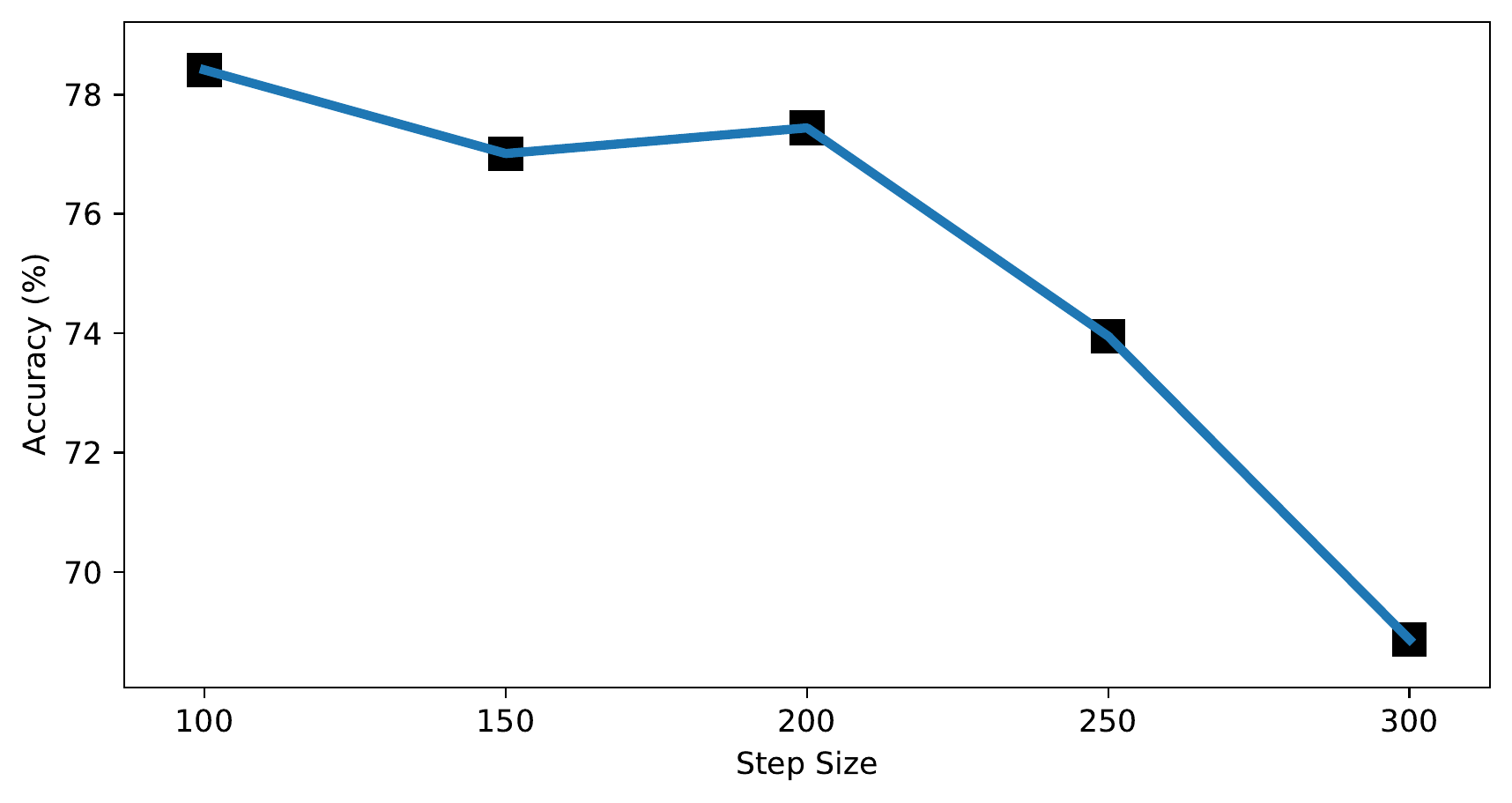}}
  \caption{Visualization of performance of SimCLR on HHAR under varying sliding window lengths and step sizes. }\label{fig:sliding-window-analysis}
\end{figure*}

\subsubsection{Cross-Person Generalization}
In previous settings, it is implicitly assumed that the training and test data distributions are well aligned. In practical scenarios of wearable-sensor-based activity recognition, however, the activity patterns of different persons inevitably vary. This is more challenging than the previous setting due to domain discrepancies among persons, and the training objective is agnostic about the test data. To deal with the domain shift, transfer learning and domain generalization approaches are developed to alleviate the domain gaps~\cite{DBLP:journals/imwut/ChangMISK20,DBLP:conf/aaai/QianPM21,DBLP:journals/corr/abs-2006-03820,DBLP:journals/ijdsn/SaputriKL14}. Here we investigate the cross-person generalization capability of the contrastive models, i.e., the model is evaluated on previously unseen target domain during test time. We follow the settings in GILE~\cite{DBLP:conf/aaai/QianPM21} to treat each person's data as a single domain. 

The experiments are conducted with leave-one-domain-out strategy, where one of the domains are chosen to be the unseen target domain and data from all the other domains are considered as training data. The UCIHAR has 30 persons in total, and we evaluate contrastive models under both data-rich scenarios where the total number of training domains is 29 and data-scarce scenarios where the number of training domains is limited to 4. For SHAR, the domains with indices $\{4, 7, 8, 10, 12, 18, 26, 27, 28, 30\}$ are removed due to incomplete classes. Then two scenarios are evaluated with the number of training domains being and 19 and 3.

The experimental results on data-rich scenarios are listed in Table.~\ref{table:cross-person-ucihar-large} and Table.~\ref{table:cross-person-shar-large} for UCIHAR and SHAR, respectively. For UCIHAR, all contrastive models show the same pattern, i.e., the performance on target domain 0 and 2 are consistently better than those on domain 1, 3 and 4. Hence, for domain 0 and 2, the domain discrepancies are alleviated. This can be caused by the fact that there exist certain source domains that share the same data characteristics with target domain 0 and 2. For SHAR, the overall accuracy is much lower than that in UCIHAR. This is reasonable since the domain gaps are larger in SHAR due to the data collection protocol. Among the 5 contrastive models, SimCLR and BYOL outperform other models on at least 2 target domains. 

The experimental results on data-scarce scenarios are shown in Table.~\ref{table:cross-person-ucihar-small} and Table.~\ref{table:cross-person-shar-small}. Compared with data-rich scenarios, the overall performance is generally lower, as it is more challenging to train models with limited training data. Interestingly, for UCIHAR, the NNCLR achieves the best performance on most target domains, which indicates that choosing nearest neighbours from distinct domains can help alleviate the mild domain shift issues in UCIHAR. However, such conclusion does not hold in SHAR, which indicates that nearest neighbours cannot handle extreme large domain discrepancies.

\begin{table}[!htbp]
  \caption{Performance of contrastive models on UCIHAR under cross-person setting with number of source domains being limited to 4.}
  \label{table:cross-person-ucihar-small}
\begin{tabular}{llllll}
\hline
\toprule
Source  & 1, 2, 3, 4 & 0, 2, 3, 4 & 0, 1, 3, 4 & 0, 1, 2, 4 & 0, 1, 2, 3 \\ \hline
Target  & 0  & 1  & 2  & 3  & 4  \\ \hline
BYOL          & 87.90      & \textbf{89.07}      & 93.26      & 79.50      & 86.09      \\
SimSiam       & 81.84      & 87.42      & 93.84      & 81.07      & 81.46      \\
SimCLR        & 96.54      & 86.75      & 96.77      & 84.23      & 91.39      \\
NNCLR     & \textbf{97.69}      & 86.42      & \textbf{97.95}     & \textbf{86.75}   & \textbf{93.38}      \\
TS-TCC        & 87.32      & 80.79      & 91.50      & 45.11      & 66.89      \\
\bottomrule
\end{tabular}
\end{table}

\begin{table}[!htbp]
  \caption{Performance of contrastive models on SHAR under cross-person setting with number of source domains being limited to 3.}
  \label{table:cross-person-shar-small}
\begin{tabular}{lllll}
\hline
\toprule
Source & 2, 3, 5 & 1, 3, 5 & 1, 2, 5 & 1, 2, 3 \\ \hline
Target & 1  & 2  & 3 & 5  \\ \hline
BYOL          & 54.17   & 39.11   & 57.57   & 40.94   \\
SimSiam       & 53.64   & 39.45   & 55.92   & 40.60   \\
SimCLR        & 52.08   & \textbf{47.34}  & \textbf{59.87}  & 40.60   \\
NNCLR         & 54.69   & 40.65   & 56.91   & 41.95   \\
TS-TCC        & \textbf{54.95}   & 36.02   & 51.97   & \textbf{42.28}  \\ 
\bottomrule
\end{tabular}
\end{table}




\subsubsection{Robustness on Wearing Diversity}
With the wider adoption of various types of wearable devices, including mobile phones, smart watches and sports bracelets, the sensor placement becomes dynamic. Wearable devices can be worn on the wrist, or be placed inside a user's trouser and shirt pockets, depending on the types of activities the user is engaged in. Consequently, it is imperative that the models should be robust to wearing diversity which pertains to the placement of the wearable sensors on the human body. More importantly, the HAR models should be able to provide accurate predictions across different wearing positions. 

Here, we follow the setups in~\cite{DBLP:journals/imwut/ChangMISK20} to evaluate contrastive models on HHAR dataset. Users in HHAR are equipped with 8 phones around waist and 4 watches worn on arms. Users then perform 6 activities: `bike', `sit', `starsdown', `stairsup', `stand' and `walk'. The results of SimCLR and TS-TCC are listed in Table~\ref{table:wearing-hhar-simclr} and Table~\ref{table:wearing-hhar-tstcc}. The results reveal that it is difficult for existing contrastive models to achieve robustness on wearing diversity. Other contrastive models have similar results, which can be found in \texttt{CL-HAR} website. 

\begin{table}
\caption{Performance of SimCLR model on HHAR under wearing diversity setting.}
\label{table:wearing-hhar-simclr}
\begin{tabular}{c c c}
\hline
\toprule
Source \textbackslash Target & phone & watch \\ \hline
phone       & 93.67 & 43.14 \\ \hline
watch       & 28.03 & 83.79 \\ \hline
phone+watch & 92.21 & 78.30 \\ 
\bottomrule
\end{tabular}
\end{table}

\begin{table}
\caption{Performance of TS-TCC model on HHAR under wearing diversity setting.}
\label{table:wearing-hhar-tstcc}
\begin{tabular}{c c c}
\hline
\toprule
Source \textbackslash Target & phone & watch \\ \hline
phone       & 91.29 & 24.27 \\ \hline
watch       & 30.64 & 78.84 \\ \hline
phone+watch & 90.36 & 73.30 \\ 
\bottomrule
\end{tabular}
\end{table}

\subsubsection{Sliding Window Matters}

So far, all contrastive models on HAR follow the default sliding window lengths and step sizes used in previous deep learning models to partition the streaming signals from wearable devices. We argue, however, that such practice implicitly violates the fully unsupervised setting of contrastive learning, since such proper sliding window lengths and step sizes are obtained under supervised setting where label information is available. For a too long or too short sliding window, the partitioned segment of data can contain multiple activities or only a small fraction of an activity. We validate this issue in Fig.~\ref{fig:sliding-window-analysis}. The results show the performance of SimCLR on HHAR with source and target domain being `watch' in Table~\ref{table:wearing-hhar-simclr}. Without any information of labels, the performance of the SimCLR model with arbitrary sliding window length and step size can vary from 68\% to 87\%. In conclusion, the implicit information leakage on sliding windows is overlooked by existing contrastive models, and more future works are desired to alleviate this issue.

\section{Conclusion}\label{sec:conclusion-discussion}

Despite the recent success of large-scale contrastive learning models, how to obtain optimal performance for small-scaled tasks remain unclear. This motivates us to provide rigorous and extensive experimental comparisons to investigate the efficacy of individual components for the task of wearable-sensor-based human activity recognition. In this paper, we evaluate state-of-the-art contrastive models for HAR from the perspectives of algorithmic level and task level. Based on our rigorous empirical observations, we show that existing contrastive models cannot handle issues in HAR perfectly, and hence there is much room for improvement in the future. In addition, we develop an open-source library \verb|CL-HAR|, which contains all experiments in our paper. The library is highly modularized, and it is handy for developing new algorithms. We hope this work can serve as a useful tool for future explorations on contrastive learning for small scaled tasks.




\newpage
\begin{figure*}[!h]
  \centering
  \includegraphics[width=0.9\linewidth]{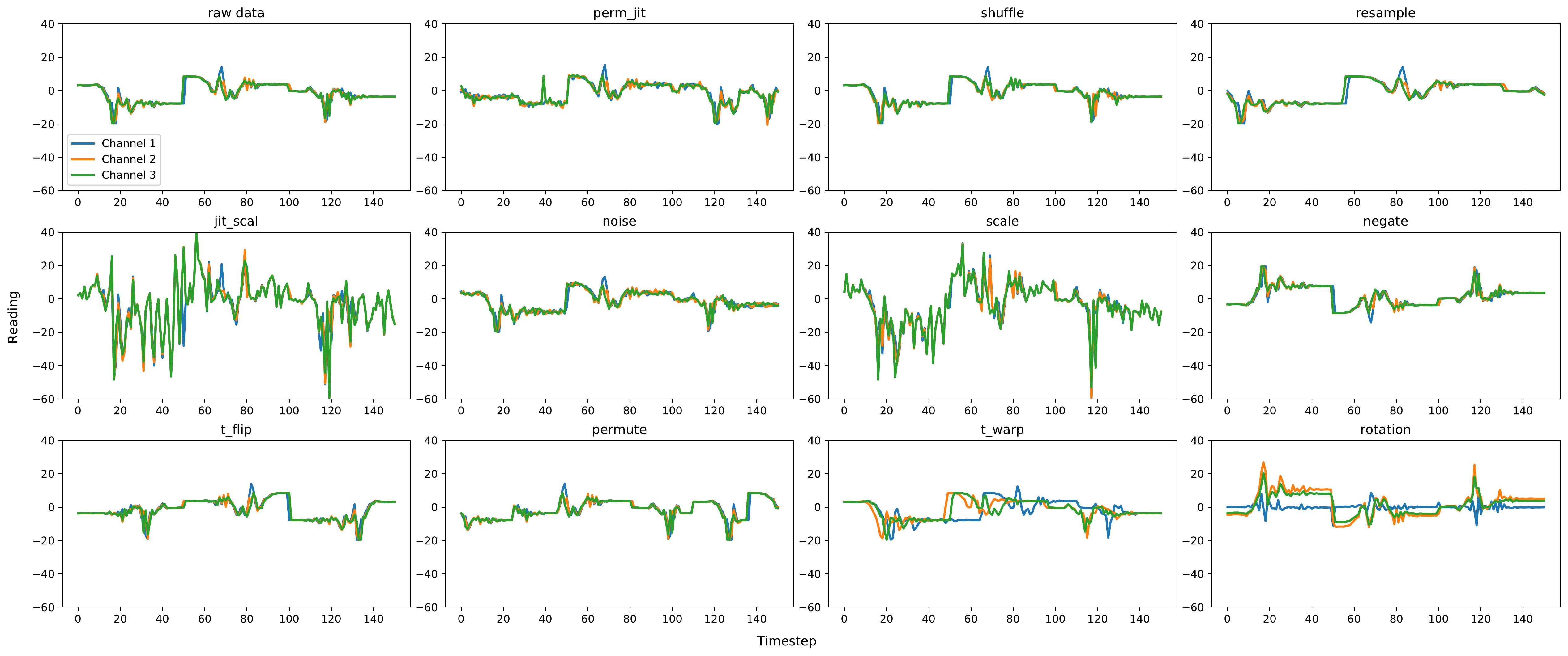}
  \caption{Visualization of time-domain augmentation transformations on activity data from SHAR.}
  \label{fig:visual-augmentation-time}
\end{figure*}

\begin{figure*}[!h]
  \centering
  \includegraphics[width=0.95\linewidth]{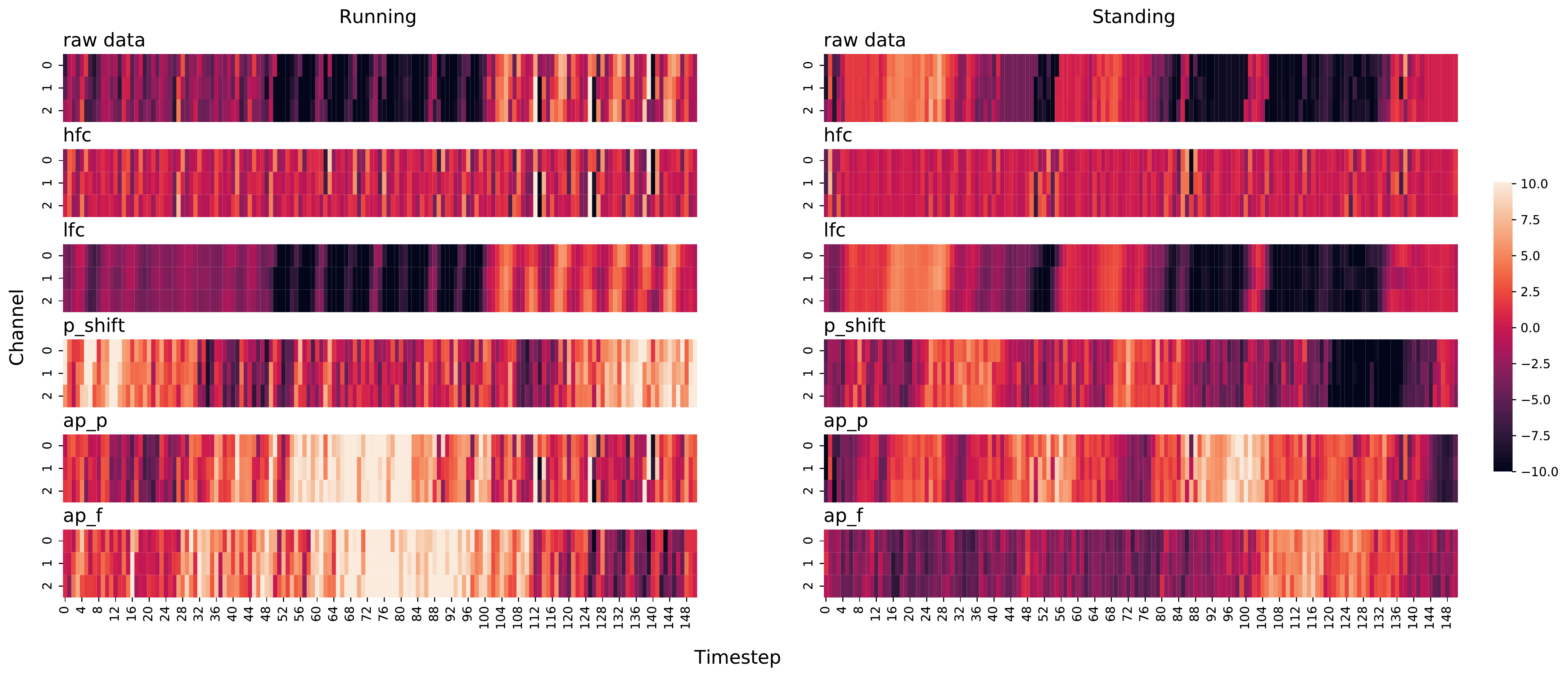}
  \caption{Visualization of frequency-domain augmentation transformations on activity data from SHAR.}
  \label{fig:visual-augmentation-freq}
\end{figure*}

\begin{table*}
\caption{Implementation Setup of Contrastive Models. lr: learning rate; bs: batch size; m: momentum; weight: weight decay; $\tau$: temperature; EMA: exponential moving average; M: memory bank size. }
\label{table:param-setup}
\begin{tabular}{cccccccccc}
\hline
\toprule
Dataset  & Model   & lr & bs & Optimizer & weight & $\tau$ & EMA & M & Epoch \\ \hline
\multirow{5}{*}{UCIHAR} & BYOL    & 5e-4          & 128   & Adam     & 1.5e-6   & -           & 0.996      & -      & 60    \\
  & SimSiam & 5e-4   & 128   & Adam        & 1e-4    & -   & -    & -   & 60    \\
   & SimCLR  & 3e-3  & 256   & Adam            & 1e-6   & 0.1    & -    & -     & 120   \\
  & NNCLR   & 3e-3   & 256    & Adam            & 1e-6   & 0.1   & -    & 1024     & 120   \\
  & TS-TCC  & 3e-4    & 128   & Adam              & 3e-4   & 0.2    & -   & -    & 40    \\ \hline
\multirow{5}{*}{SHAR} & BYOL  & 1e-3    & 64         & Adam   & 1.5e-6       & -   & 0.996      & -     & 60    \\
 & SimSiam & 3e-4   & 256    & Adam             & 1e-4     & -     & -          & -   & 60    \\
 & SimCLR  & 2.5e-3   & 256        & Adam         & 1e-6    & 0.1    & -          & -      & 120   \\
 & NNCLR   & 2e-3          & 256   & Adam          & 1e-6         & 0.1         & -          & 1024     & 120   \\
  & TS-TCC  & 3e-4    & 128   & Adam           & 3e-4   & 0.2 & -    & -     & 40    \\
\hline
\multirow{2}{*}{HHAR} & SimCLR  & 5e-3    & 256         & Adam   & 1e-6       & 0.1   & -      & -     & 120    \\
& TS-TCC  & 3e-4    & 128         & Adam   & 3e-4       & 0.2   & -      & -     & 40    \\
  \bottomrule
\end{tabular}
\end{table*}

\begin{table*}
  \caption{Implementation Details of Backbone Networks for HAR.}
  \label{table:base-encoder-imp}
  \begin{tabular}{p{0.12\textwidth} p{0.8\textwidth}}

    \toprule
Network  & Implementation Details \\
    \midrule  

DeepConvLSTM       & A 4-layer convolutional neural network, with ReLU activation after each layer, and followed by a Dropout layer and a two layer LSTM with the hidden size being 128. For each convolutional layer, the kernel size is $5\times1$ and the number of output channels is 5. The drop out rate is 0.5. \\

LSTM  & A 2-layer LSTM with hidden size of 128. \\

CNN   & A 3-layer convolutional neural network, with Batch Normalization, ReLU activation and MaxPooling after each convolutional layer and Dropout after the first convolutional layer. The kernel size of 8 and padding of size 4 are applied in the convolutional layers. The numbers of output channels are 32-64-64. The dropout rate is 0.35. \\

AE & Three-layer auto-encoder consisting of fully connected linear layers. In the encoder, the data is transformed into 8 channels by the first linear layer and flattened. The dimensions of the latter two layers are $2 \times len\_sw$ and 128 where $len\_sw$ is the length of sliding window. The decoder maps the representation to reconstruct encoder's input with a symmetric structure. \\

CAE & A convolutional auto-encoder with the above mentioned 3-layer CNN as encoder and a 3-layer de-convolution network as decoder. Each de-convolutional layer contains a max unpooling layer, a de-convolutional layer, Batch Normalization and ReLU activation.  \\

Transformer  & It consists of a linear layer and a stack of 4 identical blocks. The linear layer converts the input data to embedding vectors of 128. A token of size 128 is added to the embedded input as the representation vector. Each block is made up of a multi-head self-attention layer and a fully connected feed-forward layer. Residual connection is made around each layer. Positional encoding is added before the embedding enters the first block.  \\

\bottomrule
\end{tabular}
\end{table*}

\bibliographystyle{ACM-Reference-Format}
\bibliography{contrastive}


\begin{thebibliography}{46}


\ifx \showCODEN    \undefined \def \showCODEN     #1{\unskip}     \fi
\ifx \showDOI      \undefined \def \showDOI       #1{#1}\fi
\ifx \showISBNx    \undefined \def \showISBNx     #1{\unskip}     \fi
\ifx \showISBNxiii \undefined \def \showISBNxiii  #1{\unskip}     \fi
\ifx \showISSN     \undefined \def \showISSN      #1{\unskip}     \fi
\ifx \showLCCN     \undefined \def \showLCCN      #1{\unskip}     \fi
\ifx \shownote     \undefined \def \shownote      #1{#1}          \fi
\ifx \showarticletitle \undefined \def \showarticletitle #1{#1}   \fi
\ifx \showURL      \undefined \def \showURL       {\relax}        \fi
\providecommand\bibfield[2]{#2}
\providecommand\bibinfo[2]{#2}
\providecommand\natexlab[1]{#1}
\providecommand\showeprint[2][]{arXiv:#2}

\bibitem[Agarwal et~al\mbox{.}(2021)]%
        {DBLP:conf/iclr/AgarwalMCB21}
\bibfield{author}{\bibinfo{person}{Rishabh Agarwal}, \bibinfo{person}{Marlos~C.
  Machado}, \bibinfo{person}{Pablo~Samuel Castro}, {and}
  \bibinfo{person}{Marc~G. Bellemare}.} \bibinfo{year}{2021}\natexlab{}.
\newblock \showarticletitle{Contrastive Behavioral Similarity Embeddings for
  Generalization in Reinforcement Learning}. In
  \bibinfo{booktitle}{\emph{{ICLR}}}. \bibinfo{publisher}{OpenReview.net}.
\newblock


\bibitem[Anguita et~al\mbox{.}(2012)]%
        {DBLP:conf/iwaal/AnguitaGOPR12}
\bibfield{author}{\bibinfo{person}{Davide Anguita}, \bibinfo{person}{Alessandro
  Ghio}, \bibinfo{person}{Luca Oneto}, \bibinfo{person}{Xavier Parra}, {and}
  \bibinfo{person}{Jorge~Luis Reyes{-}Ortiz}.} \bibinfo{year}{2012}\natexlab{}.
\newblock \showarticletitle{Human Activity Recognition on Smartphones Using a
  Multiclass Hardware-Friendly Support Vector Machine}. In
  \bibinfo{booktitle}{\emph{{IWAAL}}}. \bibinfo{publisher}{Springer},
  \bibinfo{pages}{216--223}.
\newblock


\bibitem[Bracewell and Bracewell(1986)]%
        {bracewell1986fourier}
\bibfield{author}{\bibinfo{person}{Ronald~Newbold Bracewell} {and}
  \bibinfo{person}{Ronald~N Bracewell}.} \bibinfo{year}{1986}\natexlab{}.
\newblock \bibinfo{booktitle}{\emph{The Fourier transform and its
  applications}}. Vol.~\bibinfo{volume}{31999}.
\newblock \bibinfo{publisher}{McGraw-Hill New York}.
\newblock


\bibitem[Buffelli and Vandin(2020)]%
        {DBLP:journals/corr/abs-2006-03820}
\bibfield{author}{\bibinfo{person}{Davide Buffelli} {and}
  \bibinfo{person}{Fabio Vandin}.} \bibinfo{year}{2020}\natexlab{}.
\newblock \showarticletitle{Attention-Based Deep Learning Framework for Human
  Activity Recognition with User Adaptation}.
\newblock \bibinfo{journal}{\emph{CoRR}}  \bibinfo{volume}{abs/2006.03820}
  (\bibinfo{year}{2020}).
\newblock


\bibitem[Caron et~al\mbox{.}(2020)]%
        {DBLP:conf/nips/CaronMMGBJ20}
\bibfield{author}{\bibinfo{person}{Mathilde Caron}, \bibinfo{person}{Ishan
  Misra}, \bibinfo{person}{Julien Mairal}, \bibinfo{person}{Priya Goyal},
  \bibinfo{person}{Piotr Bojanowski}, {and} \bibinfo{person}{Armand Joulin}.}
  \bibinfo{year}{2020}\natexlab{}.
\newblock \showarticletitle{Unsupervised Learning of Visual Features by
  Contrasting Cluster Assignments}. In \bibinfo{booktitle}{\emph{NeurIPS}}.
\newblock


\bibitem[Chang et~al\mbox{.}(2020)]%
        {DBLP:journals/imwut/ChangMISK20}
\bibfield{author}{\bibinfo{person}{Youngjae Chang}, \bibinfo{person}{Akhil
  Mathur}, \bibinfo{person}{Anton Isopoussu}, \bibinfo{person}{Junehwa Song},
  {and} \bibinfo{person}{Fahim Kawsar}.} \bibinfo{year}{2020}\natexlab{}.
\newblock \showarticletitle{A Systematic Study of Unsupervised Domain
  Adaptation for Robust Human-Activity Recognition}.
\newblock \bibinfo{journal}{\emph{Proc. {ACM} Interact. Mob. Wearable
  Ubiquitous Technol.}} \bibinfo{volume}{4}, \bibinfo{number}{1}
  (\bibinfo{year}{2020}), \bibinfo{pages}{39:1--39:30}.
\newblock
\urldef\tempurl%
\url{https://doi.org/10.1145/3380985}
\showDOI{\tempurl}


\bibitem[Chen et~al\mbox{.}(2021)]%
        {DBLP:journals/csur/ChenZYGYL21}
\bibfield{author}{\bibinfo{person}{Kaixuan Chen}, \bibinfo{person}{Dalin
  Zhang}, \bibinfo{person}{Lina Yao}, \bibinfo{person}{Bin Guo},
  \bibinfo{person}{Zhiwen Yu}, {and} \bibinfo{person}{Yunhao Liu}.}
  \bibinfo{year}{2021}\natexlab{}.
\newblock \showarticletitle{Deep Learning for Sensor-based Human Activity
  Recognition: Overview, Challenges, and Opportunities}.
\newblock \bibinfo{journal}{\emph{{ACM} Comput. Surv.}} \bibinfo{volume}{54},
  \bibinfo{number}{4} (\bibinfo{year}{2021}), \bibinfo{pages}{77:1--77:40}.
\newblock
\urldef\tempurl%
\url{https://doi.org/10.1145/3447744}
\showDOI{\tempurl}


\bibitem[Chen et~al\mbox{.}(2020)]%
        {DBLP:conf/icml/ChenK0H20}
\bibfield{author}{\bibinfo{person}{Ting Chen}, \bibinfo{person}{Simon
  Kornblith}, \bibinfo{person}{Mohammad Norouzi}, {and}
  \bibinfo{person}{Geoffrey~E. Hinton}.} \bibinfo{year}{2020}\natexlab{}.
\newblock \showarticletitle{A Simple Framework for Contrastive Learning of
  Visual Representations}. In \bibinfo{booktitle}{\emph{{ICML}}}
  \emph{(\bibinfo{series}{Proceedings of Machine Learning Research},
  Vol.~\bibinfo{volume}{119})}. \bibinfo{publisher}{{PMLR}},
  \bibinfo{pages}{1597--1607}.
\newblock


\bibitem[Chen and He(2021)]%
        {DBLP:conf/cvpr/ChenH21}
\bibfield{author}{\bibinfo{person}{Xinlei Chen} {and} \bibinfo{person}{Kaiming
  He}.} \bibinfo{year}{2021}\natexlab{}.
\newblock \showarticletitle{Exploring Simple Siamese Representation Learning}.
  In \bibinfo{booktitle}{\emph{{CVPR}}}. \bibinfo{publisher}{Computer Vision
  Foundation / {IEEE}}, \bibinfo{pages}{15750--15758}.
\newblock


\bibitem[Dwibedi et~al\mbox{.}(2021)]%
        {DBLP:journals/corr/abs-2104-14548}
\bibfield{author}{\bibinfo{person}{Debidatta Dwibedi}, \bibinfo{person}{Yusuf
  Aytar}, \bibinfo{person}{Jonathan Tompson}, \bibinfo{person}{Pierre
  Sermanet}, {and} \bibinfo{person}{Andrew Zisserman}.}
  \bibinfo{year}{2021}\natexlab{}.
\newblock \showarticletitle{With a Little Help from My Friends:
  Nearest-Neighbor Contrastive Learning of Visual Representations}.
\newblock \bibinfo{journal}{\emph{CoRR}}  \bibinfo{volume}{abs/2104.14548}
  (\bibinfo{year}{2021}).
\newblock


\bibitem[Eldele et~al\mbox{.}(2021)]%
        {DBLP:conf/ijcai/Eldele0C000G21}
\bibfield{author}{\bibinfo{person}{Emadeldeen Eldele}, \bibinfo{person}{Mohamed
  Ragab}, \bibinfo{person}{Zhenghua Chen}, \bibinfo{person}{Min Wu},
  \bibinfo{person}{Chee~Keong Kwoh}, \bibinfo{person}{Xiaoli Li}, {and}
  \bibinfo{person}{Cuntai Guan}.} \bibinfo{year}{2021}\natexlab{}.
\newblock \showarticletitle{Time-Series Representation Learning via Temporal
  and Contextual Contrasting}. In \bibinfo{booktitle}{\emph{{IJCAI}}}.
  \bibinfo{publisher}{ijcai.org}, \bibinfo{pages}{2352--2359}.
\newblock


\bibitem[Fan et~al\mbox{.}(2021)]%
        {DBLP:journals/corr/abs-2111-01124}
\bibfield{author}{\bibinfo{person}{Lijie Fan}, \bibinfo{person}{Sijia Liu},
  \bibinfo{person}{Pin{-}Yu Chen}, \bibinfo{person}{Gaoyuan Zhang}, {and}
  \bibinfo{person}{Chuang Gan}.} \bibinfo{year}{2021}\natexlab{}.
\newblock \showarticletitle{When Does Contrastive Learning Preserve Adversarial
  Robustness from Pretraining to Finetuning?}
\newblock \bibinfo{journal}{\emph{CoRR}}  \bibinfo{volume}{abs/2111.01124}
  (\bibinfo{year}{2021}).
\newblock


\bibitem[Fang et~al\mbox{.}(2021)]%
        {DBLP:conf/iclr/FangWWZYL21}
\bibfield{author}{\bibinfo{person}{Zhiyuan Fang}, \bibinfo{person}{Jianfeng
  Wang}, \bibinfo{person}{Lijuan Wang}, \bibinfo{person}{Lei Zhang},
  \bibinfo{person}{Yezhou Yang}, {and} \bibinfo{person}{Zicheng Liu}.}
  \bibinfo{year}{2021}\natexlab{}.
\newblock \showarticletitle{{SEED:} Self-supervised Distillation For Visual
  Representation}. In \bibinfo{booktitle}{\emph{{ICLR}}}.
  \bibinfo{publisher}{OpenReview.net}.
\newblock


\bibitem[Grill et~al\mbox{.}(2020)]%
        {DBLP:conf/nips/GrillSATRBDPGAP20}
\bibfield{author}{\bibinfo{person}{Jean{-}Bastien Grill},
  \bibinfo{person}{Florian Strub}, \bibinfo{person}{Florent Altch{\'{e}}},
  \bibinfo{person}{Corentin Tallec}, \bibinfo{person}{Pierre~H. Richemond},
  \bibinfo{person}{Elena Buchatskaya}, \bibinfo{person}{Carl Doersch},
  \bibinfo{person}{Bernardo~{\'{A}}vila Pires}, \bibinfo{person}{Zhaohan Guo},
  \bibinfo{person}{Mohammad~Gheshlaghi Azar}, \bibinfo{person}{Bilal Piot},
  \bibinfo{person}{Koray Kavukcuoglu}, \bibinfo{person}{R{\'{e}}mi Munos},
  {and} \bibinfo{person}{Michal Valko}.} \bibinfo{year}{2020}\natexlab{}.
\newblock \showarticletitle{Bootstrap Your Own Latent - {A} New Approach to
  Self-Supervised Learning}. In \bibinfo{booktitle}{\emph{NeurIPS}}.
\newblock


\bibitem[Haresamudram et~al\mbox{.}(2020)]%
        {DBLP:conf/iswc/HaresamudramBAG20}
\bibfield{author}{\bibinfo{person}{Harish Haresamudram},
  \bibinfo{person}{Apoorva Beedu}, \bibinfo{person}{Varun Agrawal},
  \bibinfo{person}{Patrick~L. Grady}, \bibinfo{person}{Irfan Essa},
  \bibinfo{person}{Judy Hoffman}, {and} \bibinfo{person}{Thomas Pl{\"{o}}tz}.}
  \bibinfo{year}{2020}\natexlab{}.
\newblock \showarticletitle{Masked reconstruction based self-supervision for
  human activity recognition}. In \bibinfo{booktitle}{\emph{{ISWC}}}.
  \bibinfo{publisher}{{ACM}}, \bibinfo{pages}{45--49}.
\newblock


\bibitem[Khosla et~al\mbox{.}(2020)]%
        {DBLP:conf/nips/KhoslaTWSTIMLK20}
\bibfield{author}{\bibinfo{person}{Prannay Khosla}, \bibinfo{person}{Piotr
  Teterwak}, \bibinfo{person}{Chen Wang}, \bibinfo{person}{Aaron Sarna},
  \bibinfo{person}{Yonglong Tian}, \bibinfo{person}{Phillip Isola},
  \bibinfo{person}{Aaron Maschinot}, \bibinfo{person}{Ce Liu}, {and}
  \bibinfo{person}{Dilip Krishnan}.} \bibinfo{year}{2020}\natexlab{}.
\newblock \showarticletitle{Supervised Contrastive Learning}. In
  \bibinfo{booktitle}{\emph{NeurIPS}}.
\newblock


\bibitem[Le{-}Khac et~al\mbox{.}(2020)]%
        {DBLP:journals/access/Le-KhacHS20}
\bibfield{author}{\bibinfo{person}{Phuc~H. Le{-}Khac}, \bibinfo{person}{Graham
  Healy}, {and} \bibinfo{person}{Alan~F. Smeaton}.}
  \bibinfo{year}{2020}\natexlab{}.
\newblock \showarticletitle{Contrastive Representation Learning: {A} Framework
  and Review}.
\newblock \bibinfo{journal}{\emph{{IEEE} Access}}  \bibinfo{volume}{8}
  (\bibinfo{year}{2020}).
\newblock


\bibitem[Li et~al\mbox{.}(2021)]%
        {DBLP:conf/iclr/0001ZXH21}
\bibfield{author}{\bibinfo{person}{Junnan Li}, \bibinfo{person}{Pan Zhou},
  \bibinfo{person}{Caiming Xiong}, {and} \bibinfo{person}{Steven C.~H. Hoi}.}
  \bibinfo{year}{2021}\natexlab{}.
\newblock \showarticletitle{Prototypical Contrastive Learning of Unsupervised
  Representations}. In \bibinfo{booktitle}{\emph{9th International Conference
  on Learning Representations, {ICLR} 2021, Virtual Event, Austria, May 3-7,
  2021}}. \bibinfo{publisher}{OpenReview.net}.
\newblock
\urldef\tempurl%
\url{https://openreview.net/forum?id=KmykpuSrjcq}
\showURL{%
\tempurl}


\bibitem[Liu et~al\mbox{.}(2021)]%
        {DBLP:conf/icccn/LiuWLWYA21}
\bibfield{author}{\bibinfo{person}{Dongxin Liu}, \bibinfo{person}{Tianshi
  Wang}, \bibinfo{person}{Shengzhong Liu}, \bibinfo{person}{Ruijie Wang},
  \bibinfo{person}{Shuochao Yao}, {and} \bibinfo{person}{Tarek~F. Abdelzaher}.}
  \bibinfo{year}{2021}\natexlab{}.
\newblock \showarticletitle{Contrastive Self-Supervised Representation Learning
  for Sensing Signals from the Time-Frequency Perspective}. In
  \bibinfo{booktitle}{\emph{{ICCCN}}}. \bibinfo{publisher}{{IEEE}},
  \bibinfo{pages}{1--10}.
\newblock


\bibitem[Ma and Ghasemzadeh(2019)]%
        {DBLP:conf/aaai/MaG19}
\bibfield{author}{\bibinfo{person}{Yuchao Ma} {and} \bibinfo{person}{Hassan
  Ghasemzadeh}.} \bibinfo{year}{2019}\natexlab{}.
\newblock \showarticletitle{LabelForest: Non-Parametric Semi-Supervised
  Learning for Activity Recognition}. In \bibinfo{booktitle}{\emph{{AAAI}}}.
  \bibinfo{publisher}{{AAAI} Press}, \bibinfo{pages}{4520--4527}.
\newblock


\bibitem[Micucci et~al\mbox{.}(2016)]%
        {DBLP:journals/corr/MicucciMN16}
\bibfield{author}{\bibinfo{person}{Daniela Micucci}, \bibinfo{person}{Marco
  Mobilio}, {and} \bibinfo{person}{Paolo Napoletano}.}
  \bibinfo{year}{2016}\natexlab{}.
\newblock \showarticletitle{UniMiB {SHAR:} a new dataset for human activity
  recognition using acceleration data from smartphones}.
\newblock \bibinfo{journal}{\emph{CoRR}}  \bibinfo{volume}{abs/1611.07688}
  (\bibinfo{year}{2016}).
\newblock


\bibitem[Morales and Roggen(2016)]%
        {DBLP:journals/sensors/MoralesR16}
\bibfield{author}{\bibinfo{person}{Francisco Javier~Ord{\'{o}}{\~{n}}ez
  Morales} {and} \bibinfo{person}{Daniel Roggen}.}
  \bibinfo{year}{2016}\natexlab{}.
\newblock \showarticletitle{Deep Convolutional and {LSTM} Recurrent Neural
  Networks for Multimodal Wearable Activity Recognition}.
\newblock \bibinfo{journal}{\emph{Sensors}} \bibinfo{volume}{16},
  \bibinfo{number}{1} (\bibinfo{year}{2016}), \bibinfo{pages}{115}.
\newblock


\bibitem[Oh et~al\mbox{.}(2021)]%
        {DBLP:journals/sensors/OhALKK21}
\bibfield{author}{\bibinfo{person}{Seungmin Oh}, \bibinfo{person}{Akm
  Ashiquzzaman}, \bibinfo{person}{Dongsu Lee}, \bibinfo{person}{Yeonggwang
  Kim}, {and} \bibinfo{person}{Jinsul Kim}.} \bibinfo{year}{2021}\natexlab{}.
\newblock \showarticletitle{Study on Human Activity Recognition Using
  Semi-Supervised Active Transfer Learning}.
\newblock \bibinfo{journal}{\emph{Sensors}} \bibinfo{volume}{21},
  \bibinfo{number}{8} (\bibinfo{year}{2021}), \bibinfo{pages}{2760}.
\newblock


\bibitem[Paszke et~al\mbox{.}(2019)]%
        {DBLP:conf/nips/PaszkeGMLBCKLGA19}
\bibfield{author}{\bibinfo{person}{Adam Paszke}, \bibinfo{person}{Sam Gross},
  \bibinfo{person}{Francisco Massa}, \bibinfo{person}{Adam Lerer},
  \bibinfo{person}{James Bradbury}, \bibinfo{person}{Gregory Chanan},
  \bibinfo{person}{Trevor Killeen}, \bibinfo{person}{Zeming Lin},
  \bibinfo{person}{Natalia Gimelshein}, \bibinfo{person}{Luca Antiga},
  \bibinfo{person}{Alban Desmaison}, \bibinfo{person}{Andreas K{\"{o}}pf},
  \bibinfo{person}{Edward~Z. Yang}, \bibinfo{person}{Zachary DeVito},
  \bibinfo{person}{Martin Raison}, \bibinfo{person}{Alykhan Tejani},
  \bibinfo{person}{Sasank Chilamkurthy}, \bibinfo{person}{Benoit Steiner},
  \bibinfo{person}{Lu Fang}, \bibinfo{person}{Junjie Bai}, {and}
  \bibinfo{person}{Soumith Chintala}.} \bibinfo{year}{2019}\natexlab{}.
\newblock \showarticletitle{PyTorch: An Imperative Style, High-Performance Deep
  Learning Library}. In \bibinfo{booktitle}{\emph{NeurIPS}}.
  \bibinfo{pages}{8024--8035}.
\newblock


\bibitem[Purushwalkam and Gupta(2020)]%
        {DBLP:conf/nips/Purushwalkam020}
\bibfield{author}{\bibinfo{person}{Senthil Purushwalkam} {and}
  \bibinfo{person}{Abhinav Gupta}.} \bibinfo{year}{2020}\natexlab{}.
\newblock \showarticletitle{Demystifying Contrastive Self-Supervised Learning:
  Invariances, Augmentations and Dataset Biases}. In
  \bibinfo{booktitle}{\emph{Advances in Neural Information Processing Systems
  33: Annual Conference on Neural Information Processing Systems 2020, NeurIPS
  2020, December 6-12, 2020, virtual}}, \bibfield{editor}{\bibinfo{person}{Hugo
  Larochelle}, \bibinfo{person}{Marc'Aurelio Ranzato}, \bibinfo{person}{Raia
  Hadsell}, \bibinfo{person}{Maria{-}Florina Balcan}, {and}
  \bibinfo{person}{Hsuan{-}Tien Lin}} (Eds.).
\newblock
\urldef\tempurl%
\url{https://proceedings.neurips.cc/paper/2020/hash/22f791da07b0d8a2504c2537c560001c-Abstract.html}
\showURL{%
\tempurl}


\bibitem[Qian et~al\mbox{.}(2019b)]%
        {DBLP:conf/ijcai/QianPDM19}
\bibfield{author}{\bibinfo{person}{Hangwei Qian}, \bibinfo{person}{Sinno~Jialin
  Pan}, \bibinfo{person}{Bingshui Da}, {and} \bibinfo{person}{Chunyan Miao}.}
  \bibinfo{year}{2019}\natexlab{b}.
\newblock \showarticletitle{A Novel Distribution-Embedded Neural Network for
  Sensor-Based Activity Recognition}. In \bibinfo{booktitle}{\emph{{IJCAI}}}.
  \bibinfo{publisher}{ijcai.org}, \bibinfo{pages}{5614--5620}.
\newblock


\bibitem[Qian et~al\mbox{.}(2019a)]%
        {DBLP:conf/aaai/QianPM19}
\bibfield{author}{\bibinfo{person}{Hangwei Qian}, \bibinfo{person}{Sinno~Jialin
  Pan}, {and} \bibinfo{person}{Chunyan Miao}.}
  \bibinfo{year}{2019}\natexlab{a}.
\newblock \showarticletitle{Distribution-Based Semi-Supervised Learning for
  Activity Recognition}. In \bibinfo{booktitle}{\emph{{AAAI}}}.
  \bibinfo{publisher}{{AAAI} Press}, \bibinfo{pages}{7699--7706}.
\newblock


\bibitem[Qian et~al\mbox{.}(2021a)]%
        {DBLP:conf/aaai/QianPM21}
\bibfield{author}{\bibinfo{person}{Hangwei Qian}, \bibinfo{person}{Sinno~Jialin
  Pan}, {and} \bibinfo{person}{Chunyan Miao}.}
  \bibinfo{year}{2021}\natexlab{a}.
\newblock \showarticletitle{Latent Independent Excitation for Generalizable
  Sensor-based Cross-Person Activity Recognition}. In
  \bibinfo{booktitle}{\emph{{AAAI}}}. \bibinfo{publisher}{{AAAI} Press},
  \bibinfo{pages}{11921--11929}.
\newblock


\bibitem[Qian et~al\mbox{.}(2021b)]%
        {DBLP:journals/ai/QianPM21}
\bibfield{author}{\bibinfo{person}{Hangwei Qian}, \bibinfo{person}{Sinno~Jialin
  Pan}, {and} \bibinfo{person}{Chunyan Miao}.}
  \bibinfo{year}{2021}\natexlab{b}.
\newblock \showarticletitle{Weakly-supervised sensor-based activity
  segmentation and recognition via learning from distributions}.
\newblock \bibinfo{journal}{\emph{Artif. Intell.}}  \bibinfo{volume}{292}
  (\bibinfo{year}{2021}), \bibinfo{pages}{103429}.
\newblock
\urldef\tempurl%
\url{https://doi.org/10.1016/j.artint.2020.103429}
\showDOI{\tempurl}


\bibitem[Ramamurthy and Roy(2018)]%
        {DBLP:journals/widm/RamamurthyR18}
\bibfield{author}{\bibinfo{person}{Sreenivasan~Ramasamy Ramamurthy} {and}
  \bibinfo{person}{Nirmalya Roy}.} \bibinfo{year}{2018}\natexlab{}.
\newblock \showarticletitle{Recent trends in machine learning for human
  activity recognition - {A} survey}.
\newblock \bibinfo{journal}{\emph{WIREs Data Mining Knowl. Discov.}}
  \bibinfo{volume}{8}, \bibinfo{number}{4} (\bibinfo{year}{2018}).
\newblock


\bibitem[Saeed et~al\mbox{.}(2019)]%
        {DBLP:journals/imwut/SaeedOL19}
\bibfield{author}{\bibinfo{person}{Aaqib Saeed}, \bibinfo{person}{Tanir
  Ozcelebi}, {and} \bibinfo{person}{Johan Lukkien}.}
  \bibinfo{year}{2019}\natexlab{}.
\newblock \showarticletitle{Multi-task Self-Supervised Learning for Human
  Activity Detection}.
\newblock \bibinfo{journal}{\emph{Proc. {ACM} Interact. Mob. Wearable
  Ubiquitous Technol.}} \bibinfo{volume}{3}, \bibinfo{number}{2}
  (\bibinfo{year}{2019}), \bibinfo{pages}{61:1--61:30}.
\newblock


\bibitem[Saputri et~al\mbox{.}(2014)]%
        {DBLP:journals/ijdsn/SaputriKL14}
\bibfield{author}{\bibinfo{person}{Theresia Ratih~Dewi Saputri},
  \bibinfo{person}{Adil~Mehmood Khan}, {and} \bibinfo{person}{Seok{-}Won Lee}.}
  \bibinfo{year}{2014}\natexlab{}.
\newblock \showarticletitle{User-Independent Activity Recognition via
  Three-Stage GA-Based Feature Selection}.
\newblock \bibinfo{journal}{\emph{Int. J. Distributed Sens. Networks}}
  \bibinfo{volume}{10} (\bibinfo{year}{2014}).
\newblock


\bibitem[Shavit and Klein(2021)]%
        {DBLP:journals/access/ShavitK21}
\bibfield{author}{\bibinfo{person}{Yoli Shavit} {and} \bibinfo{person}{Itzik
  Klein}.} \bibinfo{year}{2021}\natexlab{}.
\newblock \showarticletitle{Boosting Inertial-Based Human Activity Recognition
  With Transformers}.
\newblock \bibinfo{journal}{\emph{{IEEE} Access}}  \bibinfo{volume}{9}
  (\bibinfo{year}{2021}), \bibinfo{pages}{53540--53547}.
\newblock
\urldef\tempurl%
\url{https://doi.org/10.1109/ACCESS.2021.3070646}
\showDOI{\tempurl}


\bibitem[Sheng and Huber(2019)]%
        {DBLP:conf/smc/ShengH19}
\bibfield{author}{\bibinfo{person}{Taoran Sheng} {and} \bibinfo{person}{Manfred
  Huber}.} \bibinfo{year}{2019}\natexlab{}.
\newblock \showarticletitle{Siamese Networks for Weakly Supervised Human
  Activity Recognition}. In \bibinfo{booktitle}{\emph{{SMC}}}.
  \bibinfo{publisher}{{IEEE}}, \bibinfo{pages}{4069--4075}.
\newblock


\bibitem[Sohn(2016)]%
        {DBLP:conf/nips/Sohn16}
\bibfield{author}{\bibinfo{person}{Kihyuk Sohn}.}
  \bibinfo{year}{2016}\natexlab{}.
\newblock \showarticletitle{Improved Deep Metric Learning with Multi-class
  N-pair Loss Objective}. In \bibinfo{booktitle}{\emph{{NIPS}}}.
  \bibinfo{pages}{1849--1857}.
\newblock


\bibitem[Stisen et~al\mbox{.}(2015)]%
        {DBLP:conf/sensys/StisenBBPKDSJ15}
\bibfield{author}{\bibinfo{person}{Allan Stisen}, \bibinfo{person}{Henrik
  Blunck}, \bibinfo{person}{Sourav Bhattacharya}, \bibinfo{person}{Thor~Siiger
  Prentow}, \bibinfo{person}{Mikkel~Baun Kj{\ae}rgaard},
  \bibinfo{person}{Anind~K. Dey}, \bibinfo{person}{Tobias Sonne}, {and}
  \bibinfo{person}{Mads~M{\o}ller Jensen}.} \bibinfo{year}{2015}\natexlab{}.
\newblock \showarticletitle{Smart Devices are Different: Assessing and
  MitigatingMobile Sensing Heterogeneities for Activity Recognition}. In
  \bibinfo{booktitle}{\emph{SenSys}}. \bibinfo{publisher}{{ACM}},
  \bibinfo{pages}{127--140}.
\newblock


\bibitem[Taghanaki and Etemad(2020)]%
        {DBLP:journals/corr/abs-2010-13713}
\bibfield{author}{\bibinfo{person}{Setareh~Rahimi Taghanaki} {and}
  \bibinfo{person}{Ali Etemad}.} \bibinfo{year}{2020}\natexlab{}.
\newblock \showarticletitle{Self-supervised Wearable-based Activity Recognition
  by Learning to Forecast Motion}.
\newblock \bibinfo{journal}{\emph{CoRR}}  \bibinfo{volume}{abs/2010.13713}
  (\bibinfo{year}{2020}).
\newblock


\bibitem[Tang et~al\mbox{.}(2020)]%
        {DBLP:journals/corr/abs-2011-11542}
\bibfield{author}{\bibinfo{person}{Chi~Ian Tang}, \bibinfo{person}{Ignacio
  Perez{-}Pozuelo}, \bibinfo{person}{Dimitris Spathis}, {and}
  \bibinfo{person}{Cecilia Mascolo}.} \bibinfo{year}{2020}\natexlab{}.
\newblock \showarticletitle{Exploring Contrastive Learning in Human Activity
  Recognition for Healthcare}.
\newblock \bibinfo{journal}{\emph{CoRR}}  \bibinfo{volume}{abs/2011.11542}
  (\bibinfo{year}{2020}).
\newblock


\bibitem[Tian et~al\mbox{.}(2020)]%
        {DBLP:conf/nips/Tian0PKSI20}
\bibfield{author}{\bibinfo{person}{Yonglong Tian}, \bibinfo{person}{Chen Sun},
  \bibinfo{person}{Ben Poole}, \bibinfo{person}{Dilip Krishnan},
  \bibinfo{person}{Cordelia Schmid}, {and} \bibinfo{person}{Phillip Isola}.}
  \bibinfo{year}{2020}\natexlab{}.
\newblock \showarticletitle{What Makes for Good Views for Contrastive
  Learning?}. In \bibinfo{booktitle}{\emph{NeurIPS}}.
\newblock


\bibitem[Um et~al\mbox{.}(2017)]%
        {DBLP:conf/icmi/UmPPELHFK17}
\bibfield{author}{\bibinfo{person}{Terry~Taewoong Um}, \bibinfo{person}{Franz
  Michael~Josef Pfister}, \bibinfo{person}{Daniel Pichler},
  \bibinfo{person}{Satoshi Endo}, \bibinfo{person}{Muriel Lang},
  \bibinfo{person}{Sandra Hirche}, \bibinfo{person}{Urban Fietzek}, {and}
  \bibinfo{person}{Dana Kulic}.} \bibinfo{year}{2017}\natexlab{}.
\newblock \showarticletitle{Data augmentation of wearable sensor data for
  parkinson's disease monitoring using convolutional neural networks}. In
  \bibinfo{booktitle}{\emph{{ICMI}}}. \bibinfo{publisher}{{ACM}},
  \bibinfo{pages}{216--220}.
\newblock


\bibitem[Wang et~al\mbox{.}(2019)]%
        {DBLP:journals/prl/WangCHPH19}
\bibfield{author}{\bibinfo{person}{Jindong Wang}, \bibinfo{person}{Yiqiang
  Chen}, \bibinfo{person}{Shuji Hao}, \bibinfo{person}{Xiaohui Peng}, {and}
  \bibinfo{person}{Lisha Hu}.} \bibinfo{year}{2019}\natexlab{}.
\newblock \showarticletitle{Deep learning for sensor-based activity
  recognition: {A} survey}.
\newblock \bibinfo{journal}{\emph{Pattern Recognit. Lett.}}
  \bibinfo{volume}{119} (\bibinfo{year}{2019}), \bibinfo{pages}{3--11}.
\newblock
\urldef\tempurl%
\url{https://doi.org/10.1016/j.patrec.2018.02.010}
\showDOI{\tempurl}


\bibitem[Wang et~al\mbox{.}(2021)]%
        {DBLP:journals/corr/abs-2109-02054}
\bibfield{author}{\bibinfo{person}{Jinqiang Wang}, \bibinfo{person}{Tao Zhu},
  \bibinfo{person}{Jingyuan Gan}, \bibinfo{person}{Huansheng Ning}, {and}
  \bibinfo{person}{Yaping Wan}.} \bibinfo{year}{2021}\natexlab{}.
\newblock \showarticletitle{Sensor Data Augmentation with Resampling for
  Contrastive Learning in Human Activity Recognition}.
\newblock \bibinfo{journal}{\emph{CoRR}}  \bibinfo{volume}{abs/2109.02054}
  (\bibinfo{year}{2021}).
\newblock


\bibitem[Wang and Isola(2020)]%
        {DBLP:conf/icml/0001I20}
\bibfield{author}{\bibinfo{person}{Tongzhou Wang} {and}
  \bibinfo{person}{Phillip Isola}.} \bibinfo{year}{2020}\natexlab{}.
\newblock \showarticletitle{Understanding Contrastive Representation Learning
  through Alignment and Uniformity on the Hypersphere}. In
  \bibinfo{booktitle}{\emph{{ICML}}} \emph{(\bibinfo{series}{Proceedings of
  Machine Learning Research}, Vol.~\bibinfo{volume}{119})}.
  \bibinfo{publisher}{{PMLR}}, \bibinfo{pages}{9929--9939}.
\newblock


\bibitem[Wen et~al\mbox{.}(2021)]%
        {DBLP:conf/ijcai/Wen0YSGWX21}
\bibfield{author}{\bibinfo{person}{Qingsong Wen}, \bibinfo{person}{Liang Sun},
  \bibinfo{person}{Fan Yang}, \bibinfo{person}{Xiaomin Song},
  \bibinfo{person}{Jingkun Gao}, \bibinfo{person}{Xue Wang}, {and}
  \bibinfo{person}{Huan Xu}.} \bibinfo{year}{2021}\natexlab{}.
\newblock \showarticletitle{Time Series Data Augmentation for Deep Learning:
  {A} Survey}. In \bibinfo{booktitle}{\emph{{IJCAI}}}.
  \bibinfo{publisher}{ijcai.org}, \bibinfo{pages}{4653--4660}.
\newblock


\bibitem[Yao et~al\mbox{.}(2019)]%
        {DBLP:conf/www/YaoPJZSLLLWHS0A19}
\bibfield{author}{\bibinfo{person}{Shuochao Yao}, \bibinfo{person}{Ailing
  Piao}, \bibinfo{person}{Wenjun Jiang}, \bibinfo{person}{Yiran Zhao},
  \bibinfo{person}{Huajie Shao}, \bibinfo{person}{Shengzhong Liu},
  \bibinfo{person}{Dongxin Liu}, \bibinfo{person}{Jinyang Li},
  \bibinfo{person}{Tianshi Wang}, \bibinfo{person}{Shaohan Hu},
  \bibinfo{person}{Lu Su}, \bibinfo{person}{Jiawei Han}, {and}
  \bibinfo{person}{Tarek~F. Abdelzaher}.} \bibinfo{year}{2019}\natexlab{}.
\newblock \showarticletitle{STFNets: Learning Sensing Signals from the
  Time-Frequency Perspective with Short-Time Fourier Neural Networks}. In
  \bibinfo{booktitle}{\emph{The World Wide Web Conference, {WWW} 2019, San
  Francisco, CA, USA, May 13-17, 2019}},
  \bibfield{editor}{\bibinfo{person}{Ling Liu}, \bibinfo{person}{Ryen~W.
  White}, \bibinfo{person}{Amin Mantrach}, \bibinfo{person}{Fabrizio
  Silvestri}, \bibinfo{person}{Julian~J. McAuley}, \bibinfo{person}{Ricardo
  Baeza{-}Yates}, {and} \bibinfo{person}{Leila Zia}} (Eds.).
  \bibinfo{publisher}{{ACM}}, \bibinfo{pages}{2192--2202}.
\newblock
\urldef\tempurl%
\url{https://doi.org/10.1145/3308558.3313426}
\showDOI{\tempurl}


\bibitem[Zhu et~al\mbox{.}(2021)]%
        {DBLP:journals/corr/abs-2109-01116}
\bibfield{author}{\bibinfo{person}{Yanqiao Zhu}, \bibinfo{person}{Yichen Xu},
  \bibinfo{person}{Qiang Liu}, {and} \bibinfo{person}{Shu Wu}.}
  \bibinfo{year}{2021}\natexlab{}.
\newblock \showarticletitle{An Empirical Study of Graph Contrastive Learning}.
\newblock \bibinfo{journal}{\emph{CoRR}}  \bibinfo{volume}{abs/2109.01116}
  (\bibinfo{year}{2021}).
\newblock


\end{thebibliography}

\newpage

\appendix

\section{Data Augmentation}\label{appendix:visual-augmentation}
The visualization of time-domain and frequency-domain augmentation functions are shown in Fig.~\ref{fig:visual-augmentation-time} and Fig.~\ref{fig:visual-augmentation-freq}.

\section{Experimental Configurations}\label{appendix:implementation-details}
\subsection{Datasets}

The UCIHAR dataset records 6 activities of daily living, i.e., walking, sitting, laying, standing, walking upstairs and walking downstairs. 30 participants at an age between 19 to 48 carry a smartphone on the waist while performing the activities. The smartphone records reading at a sampling rate of 50 Hz. The raw data is of 9 dimensions. 

The SHAR dataset is motivated by the observations that publicly available datasets often contain samples from subjects with too similar characteristics. Hence the dataset is designed with larger domain discrepancies in both subjects differences and activities' variations. 30 participants within an age of 18-60 perform 17 fine-grained activities, among which are 9 types of activities of daily living and 8 types of falls. The smartphones placed in the front trouser pockets of the user record 3-dimensional signals at the sampling rate of 50 Hz. 

The HHAR dataset collects 6 activities in real world scenarios, i.e., bike, sit, stairsdown, stairsup, stand, and walk. To reflect sensing heterogeneities among the readings collected from different body positions, 8 smartphones and 4 smartwatches are placed on the waists and arms of 9 participants while conducting data collection. The readings from the accelerometer and gyroscope are recorded at a frequency of 50 to 200Hz (varies across devices). Each participant performs each activity for 5 minutes. The raw data contains 6 features. 


\subsection{Implementation Details}
Due to the imbalanced nature of all the datasets, in the training stage, we set the probability of a sample being chosen in an epoch to be proportional to the inverse of the amount of the corresponding activity. Data normalization and sliding window segmentation are applied to all the datasets. For UCIHAR and SHAR datasets, the data providers have already segmented the data into sliding windows of lengths 128 (2.56 seconds) and 151 (around 3 seconds), with half sliding window length as the step size. Due to variant sampling frequencies of smart devices used in HHAR dataset, we down-sample the readings to 50Hz and apply 100 (2 seconds) and 50 as sliding window length and step size. For SHAR dataset, the samples from 10 out of 30 participants are ignored for the reason that those domains contains incomplete classes. 

In Sec.~\ref{sec:alg-level-analysis}, the datasets are randomly partitioned into training, validation and test sets, with the proportions of 64\%, 16\% and 20\%, respectively. In Sec.~\ref{sec:task-level-analysis}, Leave-One-Subject-Out evaluation method is used. The test subject or the test body part forms the test set and the rest domains are utilized as the training set. Classification accuracy on the test set is reported as the performance measure.

Unless otherwise stated, the default implementation settings of each contrastive model are listed in Table~\ref{table:param-setup}. Note that the number of required epochs is different for models, yet all models are sufficiently trained until convergence. Details of backbone networks are depicted in Table~\ref{table:base-encoder-imp}.

More implementation details and visualizations can be found in \texttt{CL-HAR} (\url{https://github.com/Tian0426/CL-HAR}).

\end{document}